\documentclass[pdflatex,sn-mathphys-ay]{sn-jnl}

\usepackage{graphicx}
\usepackage{amsmath,amssymb,amsfonts}
\usepackage{amsthm}
\usepackage{mathrsfs}
\usepackage[title]{appendix}
\usepackage{textcomp}
\usepackage{manyfoot}
\usepackage{booktabs}
\usepackage{algorithm}
\usepackage{algorithmicx}
\usepackage{algpseudocode}
\usepackage{listings}

\usepackage{subcaption}
\usepackage{siunitx}
\usepackage{url}
\usepackage{cleveref}
\usepackage{caption}
\usepackage{tabularx}
\usepackage{booktabs}

\hypersetup{
    colorlinks=true,
    linkcolor=blue,
    citecolor=blue,
    urlcolor=blue
}
\makeatletter
\AtBeginDocument{
  \def\burl#1{\url{#1}}
}
\makeatother

\theoremstyle{thmstyleone}

\theoremstyle{thmstyletwo}

\theoremstyle{thmstylethree}

\begin{document}

\title[A Fast and Accurate Fourier Neural Operator-Based Surrogate for Melt-Pool Prediction in Laser Processing]{A Fast and Accurate Fourier Neural Operator-Based Surrogate for Melt-Pool Prediction in Laser Processing}

\author[1]{\fnm{Alix} \sur{Benoit}}

\author[1]{\fnm{Toni} \sur{Ivas}}

\author[2]{\fnm{Mateusz} \sur{Papierz}}

\author[2]{\fnm{Asel} \sur{Sagingalieva}}

\author[2]{\fnm{Alexey} \sur{Melnikov}}

\author*[1]{\fnm{Elia} \sur{Iseli}}\email{elia.iseli@empa.ch}

\affil[1]{\orgdiv{Laboratory for Advanced Materials Processing}, \orgname{Swiss Federal Laboratories for Materials Science and Technology}, \orgaddress{\street{Feuerwerkerstrasse}, \city{Thun}, \postcode{3603}, \state{BE}, \country{Switzerland}}}

\affil[2]{\orgname{Terra Quantum AG}, \orgaddress{\street{Kornhausstrasse}, \city{St. Gallen}, \postcode{9000}, \state{SG}, \country{Switzerland}}}

\abstract{High-fidelity simulations of laser welding capture complex thermo-fluid phenomena, including phase change, free-surface deformation, and keyhole dynamics, but their computational cost limits large-scale process exploration and real-time use. In this work, we present the Laser Processing Fourier Neural Operator (LP-FNO), a Fourier Neural Operator (FNO)-based surrogate model that learns the parametric solution operator of laser processing from multiphysics simulations generated with FLOW-3D WELD\textsuperscript{\textregistered}. Through a reformulation of the transient problem in a reference frame moving with the laser and the application of temporal averaging, the system is recast into a quasi-steady setting suitable for operator learning, even in the stable keyhole welding regime.

The proposed LP-FNO maps process parameters to three-dimensional temperature fields and melt-pool boundaries across a broad process window spanning conduction and keyhole regimes using the non-dimensional normalized enthalpy formulation. The model achieves an average temperature relative error of approximately $2.5\%$ and intersection-over-union scores for melt-pool segmentation above $0.9$. Compared with a fully connected coordinate network, U-Net, and DeepONet, LP-FNO provides the strongest overall accuracy--efficiency trade-off for the present task, while retaining the resolution-invariant evaluation capability characteristic of FNOs. We demonstrate that an LP-FNO model trained on coarse-resolution data can be evaluated on finer grids, yielding accurate super-resolved predictions in mesh-converged conduction regimes, whereas discrepancies in keyhole regimes reflect unresolved dynamics in the coarse-mesh training data.

These results indicate that LP-FNO provides an efficient surrogate modeling framework for laser welding, enabling prediction of full three-dimensional fields and phase interfaces over wide parameter ranges in just tens of milliseconds, up to a hundred thousand times faster than traditional finite-volume multiphysics simulations.}


\maketitle
\section{Introduction}
\par 
Welding has long been a fundamental manufacturing process across many industrial sectors. Among the available techniques, laser welding has gained particular importance due to its ability to process dissimilar materials, deliver highly localized and precisely controlled heat input, operate at high speeds, and enable non-contact, easily automated processing in geometrically constrained regions. However, laser welding involves strong multi-physics effects like transient heat transfer, fluid flows in the melt-pool, phase changes, free surface deformations, to name a few.
The many influencing parameters create a strong need for fast and accurate numerical simulations to understand and optimize the quality of these processes. 
\par 
Traditionally, methods such as the Finite Volume Method (FVM) and the Finite Element Method (FEM) have been used, which provide accurate solutions and have become robust and widely used in industry \citep{michaleris_modeling_2014, mukherjee_improved_2017, mukherjee_heat_2018}. They can yield temperature fields, melt pool shapes, solidification patterns, and other characteristics which can then be validated experimentally. However, these methods can be extremely slow and computationally expensive, especially with fine meshes, strong melt-pool gradients, deep vapor depressions (keyholes), long tracks and multiple passes \citep{moreira_high-fidelity_2025}. This computational cost can become especially problematic for process control, or for situations with many parameter changes involved such as process optimization and uncertainty quantification. There exist some simplified models such as diffusion only models which can have faster run-times, but usually they have to compromise physical accuracy and can only be used for a limited set of conditions. These models frequently employ a simplified approach within the framework of thermal-only models. This method approximates the complex effects of melt pool convection by using an effective thermal conductivity \citep{de_smart_2004,saldi_effect_2013,kamara_modelling_2011},  while still suffering from the same problems as aforementioned models. Therefore there is a strong need for fast surrogate models which can retain high fidelity to the physical process of laser welding and can be used in a broad range of varying process parameters.
\par
With the recent advances in Machine Learning (ML), work has been done to fill this gap through ML based surrogate models \citep{li_fourier_2021,raissi_physics-informed_2019,karniadakis_physics-informed_2021}.
One approach has been to incorporate the underlying physics into either the architecture of neural networks, or into their loss functions, creating physics informed neural networks (PINNs). PINNs have successfully been applied to some welding and additive manufacturing (AM) problems. For example \citep{ko_review_2024} gives an overview of some of the recently published AM related work using PINNs, however it remains largely confined to the conduction (no vapor depressions) regime. Furthermore, a major drawback of standard PINNs is that they are typically trained to approximate the solution of a single PDE instance, that is for fixed parameters, geometry, boundary/initial conditions. As a result, when one changes the PDE parameters, geometry or boundary conditions, the network generally must be retrained (or at least substantially adapted). Additionally, PINNs can be difficult to optimize (balancing multiple loss terms, dealing with multi-scale phenomena, dynamic/time-dependent problems, stiff residuals) which limits their scalability to complex multi-physics or large parameter space settings \citep{li_physics-informed_2024} \citep{torres_adaptive_2025}.
\par 
Instead of trying to approximate the solution of a PDE instance directly with a neural network, one can approximate the PDE family operator which maps input functions defining the PDE family (boundary conditions, material parameters, arbitrary coefficients, etc.) to corresponding solution fields. Such a model is called a neural operator (NO). Crucially, NOs are trained on many different PDE solution fields, and can therefore generalize to new parameter sets or geometries without retraining. Currently there exist various NO architectures, such as DEEPONET, Fourier NO (FNO), Graph NO (GNO) and Convolutional NO (CNO) \citep{lu_deeponet_2021,li_fourier_2021,kovachki_neural_2023,raonic_convolutional_2023}. The strength of NOs is that once trained, they can give near instantaneous, physically consistent predictions of complete fields across wide parameter ranges. Such a model would be invaluable for applications such as laser welding, where real-time process control through a digital twin (DT), rapid process finding and optimization could strongly benefit or even only be enabled by this approach. 
\par
 There has already been some work done in this direction. In \citep{manav_meltpoolinr_2025} a neural network is presented to infer temperature field, gradients, and melt-pool shape, but only on a 2D slice. In \citep{liu_deep_2024} an example is given of an FNO based digital twin for Laser Powder Bed Fusion (L-PBF) process control. The underlying model however does not consider fluid dynamics or surface deformation. In \citep{safari_physics-informed_2025} DEEPONET and PINNs are used to simulate multi-track L-PBF processes, but only conduction is considered. And finally, \citep{yaseen_fast_2023} combined FNO and DEEPONET for modeling bead volume (volume of deposited material) and maximum melt-pool temperatures in Direct Energy Deposition (DED) processes, but likewise focuses on the conduction regime, consistent with the intended DED scope.
\par 
Taken together, these studies demonstrate the potential of machine-learning surrogates and neural operators for laser-based manufacturing, but they do not address the combination of challenges considered here. Methods restricted to two-dimensional slices cannot represent the full three-dimensional melt-pool and keyhole geometry. Thermal-only or conduction-regime models neglect melt-pool hydrodynamics, recoil-pressure-driven free-surface deformation, and the strong laser absorption changes associated with keyhole formation. Similarly, models that predict only global quantities, such as melt-pool volume or maximum melt-pool temperature, do not provide spatially resolved temperature fields or phase-interface geometries. As a result, existing approaches either reduce the dimensionality of the problem, simplify the underlying physics, or remain confined to conduction-dominated regimes. The main gap is therefore the lack of a fast surrogate model that can predict full three-dimensional fields and interfaces across both conduction and stable keyhole welding regimes while retaining the essential thermo-fluid dynamic features of high-fidelity simulations.
\par
In this work, we present the Laser Processing Fourier Neural Operator (LP-FNO), an FNO-based model for predicting melt-pool characteristics in laser processing by considering heat and fluid dynamics. Training data are generated from high-fidelity thermo-fluid simulations of single-track laser scans on Ti-6Al-4V, spanning a broad process window in laser power and scan speed. The dataset covers multiple welding regimes, ranging from conduction modes to stable keyhole welding. To enable efficient operator learning, the problem is reformulated in a reference frame moving with the laser, allowing the governing dynamics to be treated in a quasi-steady setting. Once trained, the proposed LP-FNO directly maps process parameters to three-dimensional temperature fields and melt-pool geometries in a fraction of a second, up to a hundred thousand times faster than the corresponding FVM simulation. To assess the benefit of the FNO architecture, LP-FNO is compared against several neural surrogates, including a fully connected coordinate network, a U-Net, and two DeepONet variants. The comparison shows that LP-FNO provides the most favorable accuracy--efficiency trade-off for the present dataset, particularly for melt-pool segmentation and super-resolved inference. To the best of our knowledge, this is the first surrogate modeling approach that spans conduction and keyhole welding regimes while treating keyhole dynamics within a quasi-steady operator-learning framework via time averaging. We employ a parameter sampling strategy based on normalized enthalpy \citep{hann_simple_2011},
which provides a physically motivated parameterization for covering the different welding regimes evenly.

\section{Methodology}

\subsection{Physical Model and Data Generation}
In this work Ti-6Al-4V was used as material. The introduction of non-dimensional parameter variation (see below) however was used as a first step towards model generalization not limited to a single material.

\paragraph{Simulation Software.}
\label{sec:sim_software}
The first step to build our surrogate FNO model is to choose high-fidelity simulations as ground truth to train the model. The laser melting process was simulated using a coupled multi-physics framework implemented in the commercial software FLOW-3D WELD\textsuperscript{\textregistered} 2025R1~\citep{noauthor_flow-3d_2025}. Evaporation is modeled through the added recoil pressure source term in the Navier–Stokes equations, an approach commonly used in previous studies~\citep{svenungsson2015laser,aggarwal2024unravelling} because it substantially reduces computational cost~\citep{flint_laserbeamfoam_2023} compared to fully compressible multiphase formulations~\citep{zenz2024compressible}, while retaining sufficient accuracy for predicting melt pool dynamics. FLOW-3D WELD\textsuperscript{\textregistered} solves the Navier–Stokes equations with free-surface tracking using the Volume of Fluid (VoF) method, incorporating surface tension, recoil pressure, and heat transfer with phase change. While experimental measurements could in principle be used as ground truth, this study relies exclusively on simulation data. FLOW-3D WELD\textsuperscript{\textregistered} simulation data is used as ground truth as FLOW-3D\textsuperscript{\textregistered} has already been extensively validated experimentally in both industrial and academic contexts \citep{hemmasian_surrogate_2022,ahn_novel_2025,fulco_numerical_2026}. In particular on Ti-6Al-4V LPBF processes in \cite{bayat_keyhole-induced_2019, chen_numerical_2023, ransenigo_evolution_2022, wu_parametric_2018} and in bulk Ti-6Al-4V welding in \cite{bayat_exploring_2024}. In the VoF formulation, each computational cell is assigned a metal volume fraction $\alpha$ (also referred to as metal-gas interface) representing the fraction of metal within each computational cell, with $\alpha=1$ in the metal phase, $\alpha=0$ in the gas phase, and intermediate values at the interface; this field is used as an important output to reconstruct the melt pool geometry. Thermal and fluid dynamics are coupled via temperature-dependent Marangoni convection and evaporation-induced recoil pressure, while laser energy absorption variations due to evolving keyhole geometry are accounted for using a ray-tracing approach. Thermophysical properties are taken from the FLOW-3D\textsuperscript{\textregistered} database, with the surface tension set to \SI{1.65}{J/m^2} and a temperature coefficient of \SI{-2.4e-4}{J/m^2K}~\citep{katinas2020prediction}. The computational domain is defined as a \SI{0.9}{mm} × \SI{0.4}{mm} × \SI{0.3}{mm} rectangular cuboid (length, width, height), discretized with a uniform Cartesian mesh of \SI{10}{\micro\meter} resolution. A pressure boundary condition with \SI{1}{atm} was used on the top surface, and on every other boundary, a temperature of \SI{300}{\kelvin} was imposed. Laser scanning is modeled as a Gaussian laser source of radius \SI{50}{\micro \meter} moving along the length direction. The simulation results are written at every time step of \SI{5}{\micro\second}. The total simulation time is set based on the scan speed such that the laser travels exactly \SI{0.6}{mm}. The reference change and time averaging processes are described in Section \ref{sec:refchange}. 

\paragraph{Process Parameter Variations.}
While in principle any process or material parameter could be varied, we choose to vary laser scan speed $V_{scan}$ and laser power $P$ with ranges $V_{scan} \in [0.1,1]$ \SI{}{\meter\per\second} and $P \in [40,190]$ \SI{}{\watt}, in order to cover the different melting regimes, lack of fusion (LoF), conduction melting and keyhole.
\par
\begin{figure}[h]
    \centering
    \begin{subfigure}[b]{0.48\textwidth}
        \centering
        \includegraphics[width=\linewidth]{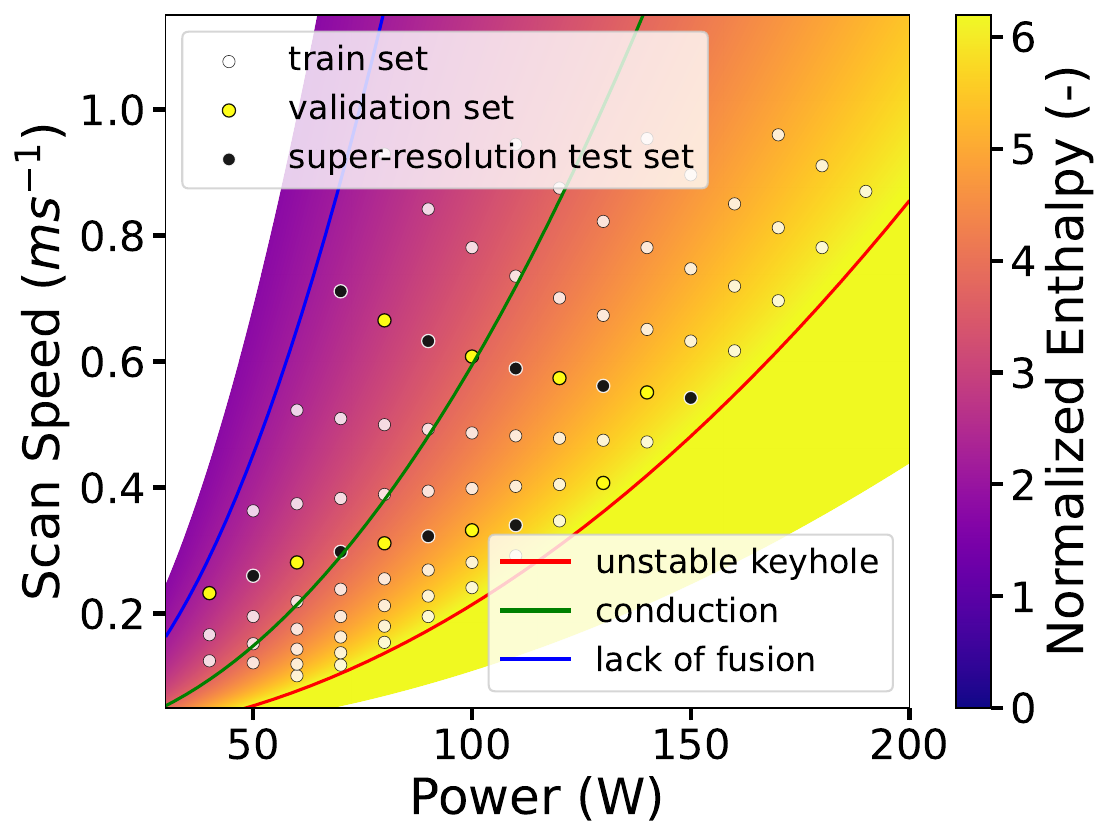}
        \caption{Power--scan speed process map colored by normalized enthalpy $H^*$}
        \label{fig:enthalpy_pv}
    \end{subfigure}
    \hfill
    \begin{subfigure}[b]{0.49\textwidth}
        \centering
        \includegraphics[width=\linewidth]{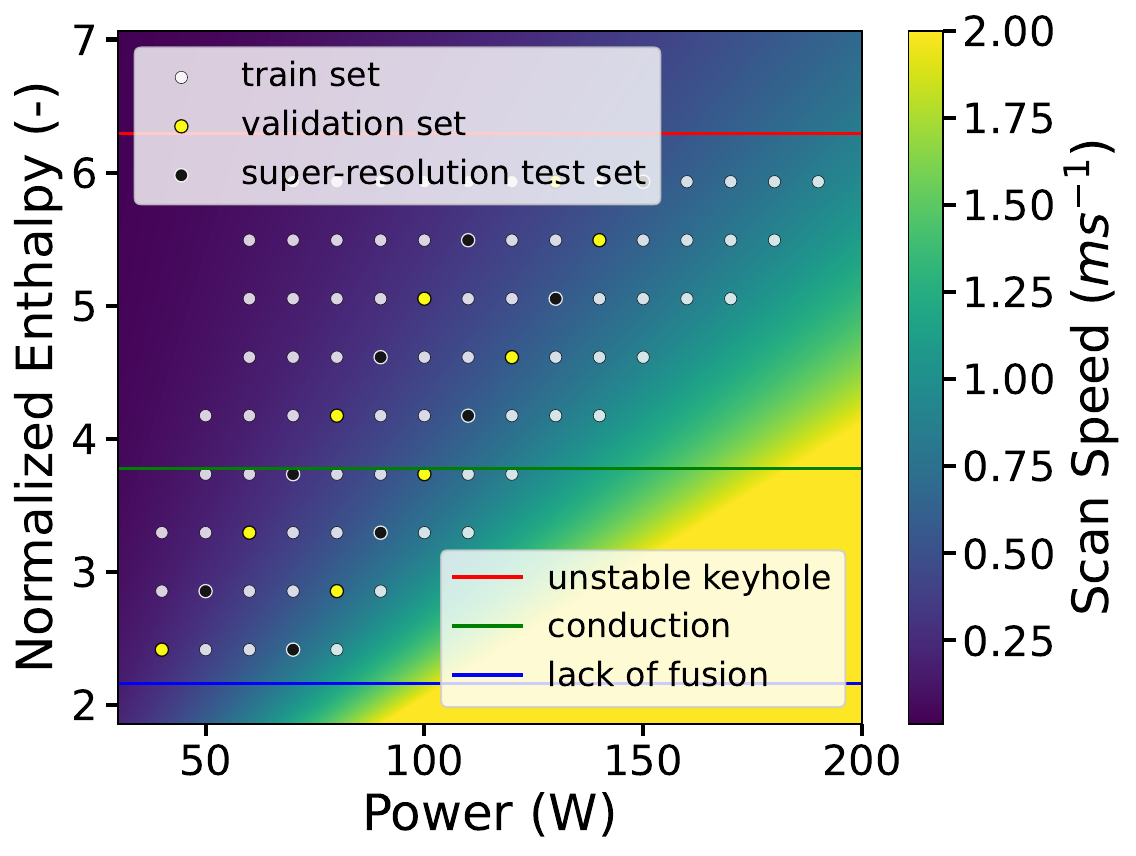}
        \caption{Power--enthalpy process map colored by scan speed $V_{\mathrm{scan}}$}
        \label{fig:enthalpy_ph}
    \end{subfigure}

    \caption{Process-space coverage of high-fidelity simulation data generated with FLOW-3D WELD. The dataset spans laser power $P\in [\SI{40}{\watt}, \SI{190}{\watt}]$ and scan speed $V_{\mathrm{scan}} \in [\SI{0.1}{\meter\per\second}, \SI{1.0}{\meter\per\second}]$, and is constructed on a grid uniform in normalized enthalpy $H^*$ and power. Markers denote the training, validation, and super-resolution test sets used for neural operator learning}
    \label{fig:enthalpy_plots}
\end{figure}

 
To decide how to sample the points within these ranges we first introduce the concept of normalized enthalpy \citep{hann_simple_2011}. Normalized enthalpy $H^*$ is a non-dimensional number that describes the keyhole formation threshold, defined in \autoref{nenthalpy} with values from \autoref{tab:nenth_params}. What is mainly important for us is that $H^* \propto \frac{P}{\sqrt{V_{scan}}}$. In order to have similar representation for both keyhole and conduction regimes, we sample our data on an equally spaced grid in $H^*$ and $P$, and derive the corresponding $V_{scan}$ from \autoref{nenthalpy}. We then select validation, and testing samples (used to evaluate super-resolution accuracy) so as to span all $H^*$ values, to make sure all regimes from LoF to keyhole are accounted for, as shown in \autoref{fig:enthalpy_plots}. An example of a conduction regime simulation is shown in \autoref{fig:conduction_coarse_fine_comparison} and a keyhole regime in \autoref{fig:keyhole_coarse_fine_comparison}.

\begin{equation}
\label{nenthalpy}
H^* \;=\; 
\frac{\eta\,P}
{\rho\,(C_p\,\Delta T_m + L)
\sqrt{\pi\,D\,\sigma^3\,V_{\text{scan}}}}
\end{equation}

\begin{table}[h]
\centering
\begin{tabular}{lll}

\hline
Symbol & Value & Description \\
\hline
$\eta$ & \SI{0.35}{} & Absorptivity \\
$\rho$ & \SI{4420}{\kg \per \meter^{3}} & Density (*) \\
$C_p$ & \SI{750}{\J\per \kg \per \K} & Specific heat capacity (**) at $T_{solidus}= \SI{1873}{K}$ \\
$\Delta T_m$ & \SI{1573}{\kelvin} & Temperature rise to melting \\
$D$ & \SI{8.1e-6}{\m^2 \per \second} & Thermal diffusivity (**) \\
$\sigma$ & \SI{50}{\micro\meter} & Beam radius (1/e$^2$) \\
$L$ & \SI{3.45e5}{\J \per \kg} & Latent Heat of Fusion \\
$P$ & \SI{40}{}--\SI{190}{\W} & Laser power range \\
$V_{\text{scan}}$ & \SI{0.1}{}--\SI{1}{\meter \per \second} & Scan speed range \\
\hline
\end{tabular}

\caption{Parameters used for normalized enthalpy evaluation. (*) Parameter taken at $T=\SI{298}{\kelvin}$. (**) Parameter taken at $T_{solidus}=\SI{1873}{K}$
}
\label{tab:nenth_params}
\end{table}

\subsection{The Fourier Neural Operator}
\par
\begin{figure}[h]
    \centering
    \includegraphics[width=\linewidth]{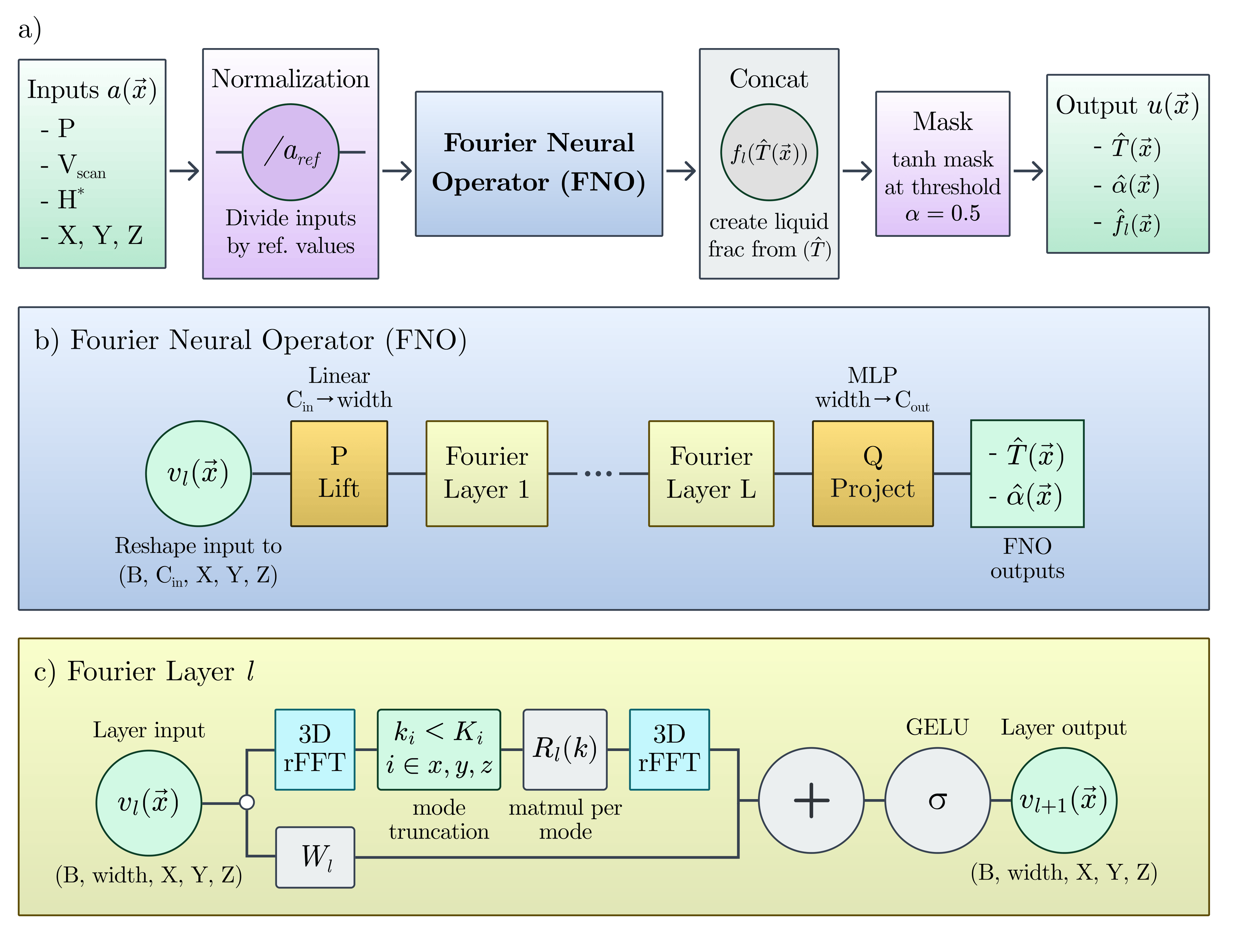}
    \caption{Architecture of LP-FNO forward pass}
    \label{fig:LPFNO-diagram}
\end{figure}

The classical Fourier Neural Operator architecture is used, which is described for modeling the laser melting, in the following section. One can see an overview diagram of the LP-FNO in \autoref{fig:LPFNO-diagram}.
\paragraph{Mathematical Setting.}
A class of steady (time-independent) partial differential equations posed on a bounded domain $D \subset \mathbb{R}^d$ is assumed. The problem is written in operator form as follows:
\begin{equation}
\begin{cases}
\mathcal{P}(u, a) = 0, & \text{in } D, \\
u = g, & \text{on } \partial D,
\end{cases}
\end{equation}
where $u \in \mathcal{U}$ denotes the unknown solution field and $a \in \mathcal{A}$ represents a set of input parameters, typically corresponding to spatially varying coefficients, source terms, or material properties of the PDE. In our case $u$ is the concatenation of temperature field $T$, the volume fraction $\alpha$, and the liquid fraction $f_\ell(T)$ (computed from $T$, defined in Section \ref{sec:refchange}) and the variable parameter field is $a = (P, V_{\mathrm{scan}}, H^*)$, which is constant in space. The boundary conditions $g$ are described in the previous section.
\par
The operator $\mathcal{P} : \mathcal{U} \times \mathcal{A} \rightarrow \mathcal{F}$ denotes the differential operator defining the PDE, with $\mathcal{U}$, $\mathcal{A}$, and $\mathcal{F}$ taken as Banach spaces equipped with appropriate norms. Under standard well-posedness assumptions, the PDE admits a unique solution for each admissible parameter field $a$.
\paragraph{Solution Operator.}
This setting induces a solution operator
\begin{equation}
\mathcal{G}^{\dagger} : \mathcal{A} \rightarrow \mathcal{U}, \quad a \mapsto u,
\end{equation}
which maps the PDE parameters directly to the corresponding solution. The goal of operator learning is to approximate $\mathcal{G}^{\dagger}$ from data, rather than learning a solution for a fixed instance of $a$.

\paragraph{For our case.}
In this work, the governing equations are the Navier-Stokes equations, and the heat equation with added laser source in the laser frame of reference, along with the VoF equation \citep{flint_laserbeamfoam_2023}.
\begin{equation}
\frac{\partial \alpha}{\partial t} + \nabla \cdot (\alpha \mathbf{U}) = 0,
\end{equation}
Where $\alpha$ is the volume fraction and $U$ is the velocity field (not explicitly computed in our case).
We look for stationary solutions of the equations, as part of our quasi-steady state assumption.

The FNO is employed to learn an approximation of the solution operator $\mathcal{G}^{\dagger}$ that is resolution-invariant and defined directly at the operator level, enabling generalization across different resolutions of the domain discretization. The exact description of the governing equations is not necessary as they are learned implicitly by the neural operator through the parameter-solution pairs during training.

\paragraph{Discretized operator learning problem.}
In practice, the parameter field $a(\mathbf{x})$ and the corresponding solution field $u(\mathbf{x})$ are observed on a grid (the FNO can be extended to non-cartesian meshes \citep{li_fourier_2023}, but we still do the classical formulation here). Let $D_h$ denote a discretization of $D$ with $n$ points (e.g., $n = n_x n_y n_z$ for a structured mesh in 3D), then the training dataset is given by pairs:
\begin{equation}
\left\{ \big(a^{(j)}|_{D_h},\, u^{(j)}|_{D_h}\big) \right\}_{j=1}^{N},
\end{equation}
where each pair corresponds to one quasi-steady simulation instance. The learning task is to construct a parametric operator $\mathcal{G}_{\theta}$ such that $\mathcal{G}_{\theta}(a) \approx \mathcal{G}^{\dagger}(a)$, and such that the approximation remains meaningful when evaluated on a different discretization $D_{h'}$ (i.e., generalization across resolutions).

\paragraph{Architecture overview.}
The FNO approximates $\mathcal{G}^{\dagger}$ using a composition of $L$ operator layers acting on a latent field. The construction is typically expressed as
\begin{equation}
\mathcal{G}_{\theta} = \mathcal{Q} \circ \mathcal{K}^{(L)} \circ \cdots \circ \mathcal{K}^{(1)} \circ \mathcal{P},
\end{equation}
where $\mathcal{P}$ denotes an input ``lifting'' map into a higher-dimensional channel space, $\mathcal{K}^{(\ell)}$ are Fourier layers, and $\mathcal{Q}$ is an output projection back to the physical quantity of interest. An initial latent representation is then formed by
\begin{equation}
v_0(\mathbf{x}) = \mathcal{P}(a(\mathbf{x}), \mathbf{x}),
\end{equation}
where $\mathbf{x}=(x,y,z)$ denotes the spatial coordinate.

\paragraph{Fourier layer as global convolution.}
Each Fourier layer is designed to mix information over the entire domain efficiently, which is important for problems where long-range coupling is present. The update at layer $\ell$ is written point-wise in physical space as

\begin{equation}
v_{\ell+1}(\mathbf{x}) = \sigma\!\left( \big(\mathcal{W}_{\ell} v_{\ell}\big)(\mathbf{x}) + \big(\mathcal{F}^{-1} \circ \mathcal{R}_{\ell} \circ \mathcal{F}\big)\!\left[v_{\ell}\right](\mathbf{x}) \right),
\label{eq:fno_layer}
\end{equation}

where $\sigma$ denotes a nonlinearity (GELU \citep{hendrycks_gaussian_2016} in our case), $\mathcal{W}_{\ell}$ is a learned point-wise linear map, and the second term implements a learned integral operator through the Fourier transform $\mathcal{F}$. This second term corresponds to a global convolution with a translation-invariant kernel; in Fourier space, such a convolution becomes multiplication by a mode-dependent operator.

\paragraph{Mode truncation and learnable spectral weights.}
For a discretized field $v_{\ell}$, the Fourier transform yields coefficients $\widehat{v}_{\ell}(\boldsymbol{k})$ indexed by wave-vectors $\boldsymbol{k}$. In practice, the Fourier-space operator $\mathcal{R}_{\ell}$ is implemented as a block-diagonal operator, where each retained Fourier mode $\boldsymbol{k}$ is associated with a learnable matrix $\mathbf{R}_{\ell}(\boldsymbol{k})$ acting on the channel dimension. In the FNO, only a fixed set of low-frequency modes is retained:
\begin{equation}
\widehat{v}_{\ell+1}(\boldsymbol{k}) =
\begin{cases}
\mathbf{R}_{\ell}(\boldsymbol{k}) \, \widehat{v}_{\ell}(\boldsymbol{k}), & \boldsymbol{k} \in \mathcal{K}_{m}, \\
0, & \text{otherwise},
\end{cases}
\end{equation}
where $\mathcal{K}_{m}$ denotes the retained index set (e.g.\ the first $m_1 \times \cdots \times m_d$ modes). The truncation reflects the empirical observation that many PDE solution operators are well-approximated by their low-frequency content, while higher frequencies are either weakly excited or dominated by discretization noise. Computationally, the cost per layer scales as $\mathcal{O}(n \log n)$ due to the FFT, while the learnable spectral multiplication acts only on $\mathcal{O}(m_1 \cdots m_d)$ modes.

\paragraph{Input/output maps and physical fields.}
The lifting map $\mathcal{P}$ and projection map $\mathcal{Q}$ are typically implemented as pointwise multilayer perceptrons (MLPs). This design keeps the network compatible with variable grid sizes, which means that the same parameters are reused at each spatial location, and the Fourier mixing step is defined through FFTs on the current grid. For multi-physics settings, multiple input channels (e.g.\ power, scan speed, normalized enthalpy, and grid coordinates in our case) may be concatenated into $a(\mathbf{x})$, while multiple output channels may be predicted jointly (e.g. temperature, volume fraction, and liquid fraction in our case).

\paragraph{Handling of boundaries.}
In the current implementation, boundary conditions are not imposed as explicit hard constraints inside the FNO; instead, their effect is learned from simulation data generated with those boundary conditions. The FNO receives normalized coordinate channels together with process parameters, so the network can learn location-dependent behavior near domain boundaries. No separate boundary-loss term, such as a Dirichlet boundary, Neumann boundary, or surface-specific boundary condition is imposed during the forward pass. 
The configured zero padding extends the computational grid with a halo of zeros before the spectral convolution/Fourier transform. This padding is used to reduce the artificial periodic wrap-around inherent to Fourier transforms: without padding, opposite sides of the physical domain are implicitly adjacent in spectral space, whereas the padded halo provides a buffer region so learned global convolutions are less likely to couple unrelated boundaries directly. The padded values should therefore be interpreted as a numerical device for the spectral operator, not as a physical zero boundary condition; predictions are learned on the original physical domain. The top surface pressure boundary is not treated differently from fixed-temperature boundaries in the present FNO model. In fact, for the current temperature/phase runs, pressure boundary conditions enter only indirectly through the training data produced by the CFD solver; the FNO itself predicts T and $\alpha$ and applies the same spectral architecture and padding uniformly across all spatial boundaries.

\paragraph{Training objective.}
Training is performed using supervised pairs $(a^{(j)}, u^{(j)})$ by minimizing the empirical risk function:
\begin{equation}
\min_{\theta}\; \frac{1}{N}\sum_{j=1}^{N} \left\| \mathcal{G}_{\theta}\!\left(a^{(j)}\right) - u^{(j)} \right\|_{\mathcal{U}_h}^{2},
\end{equation}
where $\|\cdot\|_{\mathcal{U}_h}$ denotes a discretized norm (in our case $L^2$ norm over grid points). When multiple fields are predicted, there are multiple ways to aggregate them, the most simple being a weighted sum. We opt for the ReLoBRaLo \citep{bischof_multi-objective_2025} aggregator, which was specifically designed for PINNs.

\paragraph{Resolution invariance in practice.}
\label{sec:super-resolution}
The resolution-invariant behavior of the FNO is a consequence of (i) the point wise maps $\mathcal{P}$ and $\mathcal{Q}$ that do not depend on grid size and (ii) the Fourier mixing defined by a fixed set of retained modes $(m_1,\ldots,m_d)$ rather than a fixed number of grid points. As a result, the learned spectral weights $\mathbf{R}_{\ell}(\boldsymbol{k})$ for a specific mesh resolution can be applied on another mesh resolution, provided the retained Fourier indices are representable on the evaluation grid. For non-periodic domains or non-periodic boundary conditions, padding or other boundary treatments are applied before computing the FFT so that the spectral convolution remains numerically stable. In our case, while we have a periodic domain on some sides, most are equal to $\SI{300}{\kelvin}$ for the temperature field, the top side breaks periodicity; we thus employ a zero padding of size 9 around our domain boundary.  

\subsection{Other tested Architectures}
\label{sec:baseline_architectures}

To assess the contribution of the FNO architecture, we compare LP-FNO against four additional neural surrogate architectures: a fully connected coordinate neural network, a U-Net \citep{ronneberger_u-net_2015}, and two DeepONet \citep{lu_deeponet_2021} variants with either a fully connected or Fourier-feature trunk.

\paragraph{Fully Connected Coordinate-based Neural Network.}
The fully connected neural network (FC-NN) treats the surrogate modeling problem as a pointwise regression task. Each spatial location $x = (x,y,z)$ is converted into an independent training sample by concatenating its normalized coordinates with the global process parameters, such as laser power $P$, scan speed $V_{\mathrm{scan}}$, and normalized enthalpy $H^\ast$. The resulting input vector is passed through a multilayer perceptron with skip connections to predict the target fields at that spatial point,
\begin{equation}
    \hat{u}(x) = \mathcal{N}_{\theta}\left(x, P, V_{\mathrm{scan}}, H^\ast\right),
\end{equation}
where $\hat{u}(x)$ denotes the predicted field values, including temperature and phase-related quantities. This model provides a direct measure of how much of the solution structure can be learned from local coordinates and process parameters alone. Because the FC-NN is evaluated pointwise, the prediction at each spatial location depends only on the local coordinates and global process parameters provided as input. Spatial coherence is therefore learned only implicitly through the shared network parameters and the supervised loss over many sampled points, rather than being enforced explicitly through convolutional neighborhoods, spectral mixing, or other field-level coupling mechanisms.

\paragraph{U-Net.}
The U-Net architecture retains the data in structured grid form and learns a direct field-to-field mapping using a convolutional encoder--decoder architecture \cite{ronneberger_u-net_2015}. In contrast to the pointwise FC-NN, the U-Net processes the full spatial domain simultaneously and can exploit local spatial correlations through convolutional filters. The encoder progressively compresses the grid representation into increasingly abstract latent features, while the decoder reconstructs the output fields at the original resolution. Skip connections between corresponding encoder and decoder levels preserve fine-scale spatial information during reconstruction.  
Nevertheless, the U-Net remains fundamentally grid-based: its learned representation is tied to the discretization used during training and does not provide the same resolution-invariant operator formulation as LP-FNO.

\paragraph{DeepONet.}
The DeepONet architectures use an operator-learning architecture with separate branch and trunk networks \cite{lu_deeponet_2021}. The branch network encodes non-coordinate conditioning information, here given by the process parameters, while the trunk network encodes the spatial coordinate at which the solution is queried. Their latent representations are combined to evaluate the predicted solution field pointwise,
\begin{equation}
    \hat{u}(x; a) =
    \sum_{k=1}^{p}
    b_k(a) \, t_k(x),
\end{equation}
where $(a = (P, V_{\mathrm{scan}}, H^\ast))$, $(b_k(a))$ are the branch-network outputs, and $(t_k(x))$ are the trunk-network outputs. In this sense, DeepONet represents a mapping from input conditions to a continuous spatial field, rather than predicting an entire grid through convolutional or spectral layers. This provides a useful comparison with LP-FNO, since both architectures aim to learn an operator, but DeepONet evaluates the solution pointwise through a branch--trunk factorization.

Two DeepONet variants are considered. In the fully connected trunk variant (DPN-FCT), the coordinate trunk is implemented as a standard multilayer perceptron. This tests whether the DeepONet factorization itself is beneficial for the present laser-processing surrogate task without imposing an explicit spectral coordinate basis. In the Fourier-trunk variant (DPN-FourierT), the coordinate input is first transformed using Fourier feature embeddings before being passed through the trunk network. This enriches the coordinate representation with periodic basis functions and can improve the representation of high-frequency or rapidly varying spatial structure. In both cases, the branch network remains responsible for encoding the global process condition, while the trunk network determines how the learned operator is evaluated over space.

These architectures therefore span three distinct modeling strategies. The FC-NN tests purely pointwise coordinate-conditioned regression, the U-Net tests convolutional grid-based field prediction, and the DeepONet variants test pointwise operator learning with either learned or Fourier-enriched coordinate bases. Their comparison with LP-FNO isolates the benefit of using spectral operator layers that act on the full field and mix information globally in Fourier space.

\paragraph{Super-resolution capabilities of the different architectures.}
The theoretical ability to perform super-resolution differs across these architectures. The U-Net is the most directly tied to the training discretization, since its convolutional encoder--decoder structure maps one fixed grid representation to another and does not define an intrinsic continuous or resolution-invariant operator. By contrast, the FC-NN and DeepONet architectures can, in practice, be evaluated at arbitrary spatial coordinates and can therefore produce predictions on a finer grid than the one used during training. For the FC-NN, this corresponds to coordinate-based regression: each queried point is evaluated using its coordinates and the global process parameters, while spatial coherence is learned only implicitly through the shared network weights rather than through explicit convolutional, spectral, or field-level interactions. For DeepONet, this corresponds to pointwise evaluation of a learned operator through the branch--trunk factorization: the branch network encodes the process parameters, while the trunk network evaluates the output field at queried spatial coordinates. Thus, DeepONet does provide a form of discretization-independent operator evaluation, but it does not process the full output grid through explicit field-level convolutional or spectral interactions during inference. The FNO super-resolution mechanism is different: the same learned spectral operator, parameterized by retained Fourier modes rather than by grid points, is applied to the full field on the new discretization. This provides a direct form of resolution-invariant field-level operator evaluation, although its physical fidelity remains limited by the resolution and convergence of the training data.

\subsection{Case specific processing}

In order to use the FNO architecture for laser melting, different adjustments were made for consistent training. 

\paragraph{Reference frame change.}
\label{sec:refchange}
To transform the data into the reference frame of a laser moving at a constant speed $V_{scan}$, the temporal sampling is chosen such that the laser displacement aligns exactly with the underlying uniform computational grid. Given a spatial grid with uniform spacing $\Delta x$, the time increment $\Delta t$ is therefore selected so that the laser travels precisely one grid cell per time step,
\begin{equation}
\Delta t = \frac{\Delta x}{V_{scan}}.
\end{equation}
With this choice, the laser position coincides with grid points at every time step in the moving reference frame, eliminating the need for spatial interpolation. Consequently, if the laser traverses a total distance corresponding to $m$ grid cells, only $m$ discrete time steps are retained in the transformed frame, yielding a total elapsed time of $m\,\Delta x / V$. This approach aligns with treating the laser’s travel distance as the effective evolution parameter, since simulations are run until a specified distance is reached rather than for a fixed physical time. We subsequently refer to time steps in the laser reference frame through indices as $t_k = k\Delta t$, $k\in \mathbb{N}$.

\paragraph{Quasi-steadiness assumption and time averaging.}
\label{time_avging}
In the reference frame moving with the laser at constant speed, temporal variations in the temperature field and melt-pool interface were largely confined to small-scale interface fluctuations, with minimal contribution from large-scale melt-pool evolution. To evaluate whether these fields could be treated as quasi-static, we analyzed the temporal differences between consecutive snapshots in the moving frame. This procedure has in particular limitations for deep, highly fluctuating keyhole conditions, which are characterized by large aspect-ratio vapor cavities, pronounced humping of the keyhole walls, possible periodic keyhole collapse, and the entrainment of gas pockets. These conditions lead to melt-pool dynamics that are fundamentally transient and stochastic. Under such conditions, the keyhole geometry fluctuates on timescales comparable to or shorter than the averaging window, meaning that temporal averaging no longer isolates a slowly varying mean field but instead blurs physically meaningful transient events that govern defect formation. In practical laser‑processing applications, however, such highly fluctuating regimes are typically avoided through parameter selection and process control. This makes the more stable operating conditions—those captured by the model in this work—the primary domain of practical relevance.

\begin{figure}[h]
    \centering
    \begin{subfigure}[t]{0.48\linewidth}
        \centering
        \includegraphics[width=\linewidth]{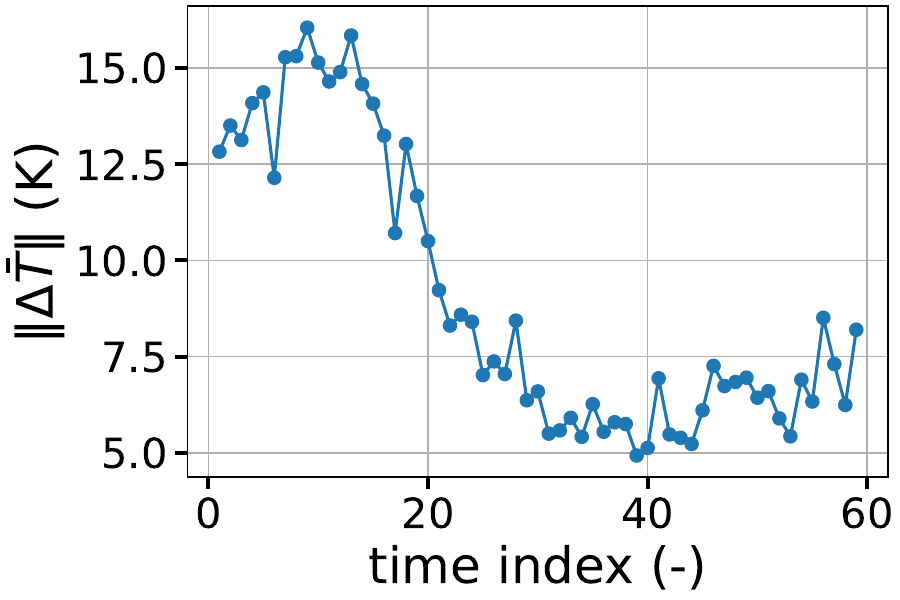}
        \caption{No temporal averaging}
        \label{fig:time_diff_no_avg_conduction}
    \end{subfigure}
    \hfill
    \begin{subfigure}[t]{0.48\linewidth}
        \centering
        \includegraphics[width=\linewidth]{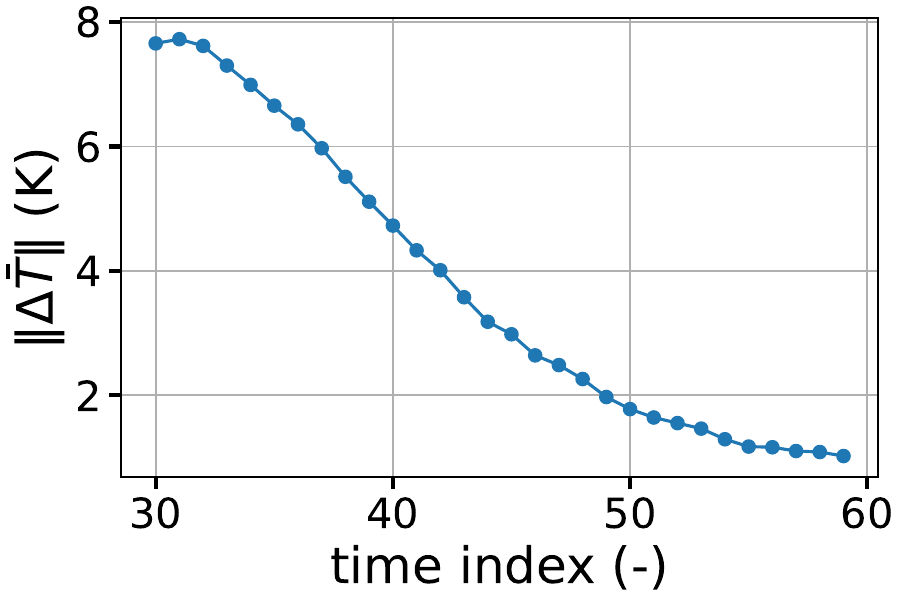}
        \caption{Averaging over a 30-step window}
        \label{fig:time_diff_avg_conduction}
    \end{subfigure}
    \caption{Conduction regime ($P=\SI{70}{\watt}$, $V_{scan}=\SI{0.298}{\meter\per\second}$). Temporal evolution of the mean absolute temperature difference in the laser reference frame.
    Quasi-static behavior appears around index $t_k=t_{30}$ for non-averaged data, and in averaged data from $t_k=t_{50}$. Time averaging is defined in \autoref{eq:window_avg}}
    \label{fig:temporal_differences_conduction}
\end{figure}

\begin{figure}[h]
    \centering
    \begin{subfigure}[t]{0.48\linewidth}
        \centering
        \includegraphics[width=\linewidth]{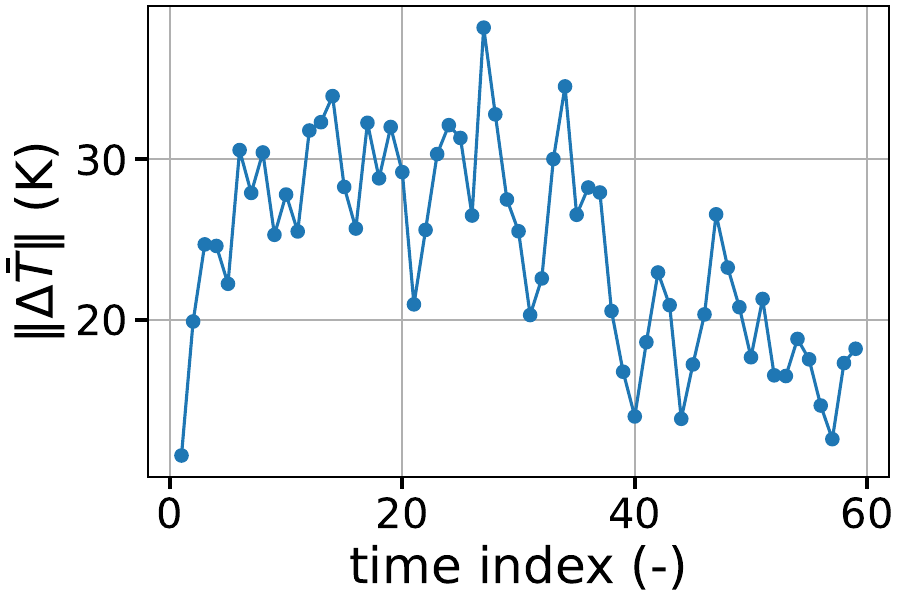}
        \caption{No temporal averaging}
        \label{fig:time_diff_no_avg_keyhole}
    \end{subfigure}
    \hfill
    \begin{subfigure}[t]{0.48\linewidth}
        \centering
        \includegraphics[width=\linewidth]{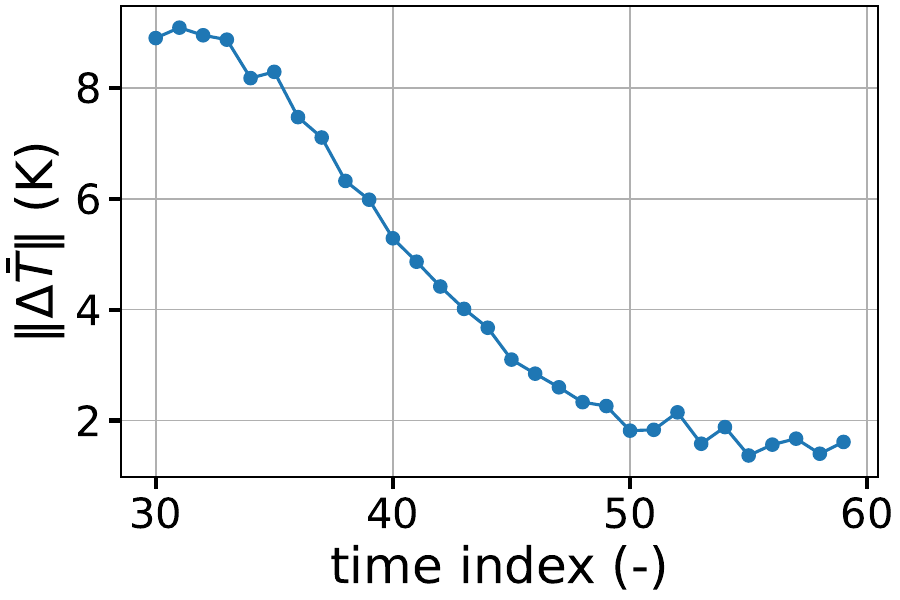}
        \caption{Averaging over a 30-step window}
        \label{fig:time_diff_avg_keyhole}
    \end{subfigure}
    \caption{Keyhole regime ($P=\SI{150}{\watt}$, $V_{scan}=\SI{0.542}{\meter\per\second}$). Temporal evolution of the mean absolute temperature difference in the laser reference frame.
    Temporal averaging suppresses high-frequency interface fluctuations,
    revealing quasi-static behavior once the laser has traveled a sufficient distance. Time averaging is defined in \autoref{eq:window_avg}}
    \label{fig:temporal_differences_keyhole}
\end{figure}

Figures~\ref{fig:temporal_differences_conduction} and~\ref{fig:temporal_differences_keyhole} show the mean absolute difference between the temperature field at two successive time steps, $\|\Delta T(t_{k})\| = \|T(t_k) - T(t_{k-1})\|$, for both conduction and keyhole welding regimes. Without temporal averaging, significant fluctuations persist in the keyhole regime (\autoref{fig:time_diff_no_avg_keyhole}), reflecting the inherently unsteady dynamics of the vapor cavity and the associated melt-pool interface. In contrast, the conduction regime exhibits substantially smaller variations, indicating a closer proximity to steady behavior.

To isolate the large-scale, slowly varying structure of the melt pool, a temporal averaging procedure is applied over a sliding window, defined as 
\begin{equation}
    \label{eq:window_avg}
    \bar T(t_k) = \frac{1}{n}\sum_{i=k-n}^k T(t_i),
\end{equation}
with $n=30$ time steps in the moving frame. After averaging, the magnitude of $\|\Delta \bar T(t_k)\|$ approaches zero in both regimes. This demonstrates that, once high-frequency interface oscillations are filtered out, the temperature field and melt-pool geometry exhibit only weak residual temporal dependence after the laser has traveled a sufficient distance.

These observations justify treating the melt pool as quasi-static in the laser reference frame, provided that an appropriate temporal averaging is employed.

\paragraph{Normalization of fields}
All fields used by the LP-FNO were normalized by dividing by a typical reference value, so that they are all on the same order of magnitude to make learning easier. The used reference values are indicated in \autoref{tab:reference_scales}.
\begin{table*}[h]
\centering
\caption{Reference values used for normalization}
\label{tab:reference_scales}
\begin{tabular}{lll}
\hline
\textbf{Characteristic parameter} & \textbf{Symbol} & \textbf{Reference value} \\
\hline
Length        & $L$    & \SI{1e-4}{\meter} \\
Temperature   & $T$    & \SI{3000}{\kelvin} \\
Velocity      & $V$    & \SI{1e-1}{\meter\per\second} \\
Power         & $P$    & \SI{10}{\watt} \\
Normalized enthalpy & $H$    & \num{5.9} \\
\hline
\end{tabular}
\end{table*}

\paragraph{Computation of liquid fraction.}
The liquid fraction $f_\ell$ is the amount of liquid compared to solid and is defined as follows:
\begin{equation}
\label{eq:fl}
f_\ell(T(\mathbf{x}))=
\begin{cases}
0, & T(\mathbf{x}) \le T_s,\\[4pt]
\dfrac{T(\mathbf{x})-T_s}{T_l-T_s}, & T_s < T(\mathbf{x}) < T_l,\\[6pt]
1, & T(\mathbf{x}) \ge T_l.
\end{cases}
\end{equation}
where $T_s$ and $T_l$ are the solidus and liquidus temperatures of the metal respectively. For Ti-6Al-4V, $T_s = \SI{1873}{\kelvin}$ and $T_l= \SI{1923}{\kelvin}$ were used, based on the FLOW-3D\textsuperscript{\textregistered} database. The region where $0 < f_\ell < 1$ is called the "mushy zone" \citep{kurz_fundamentals_2023}. In this regime,  the solid dendrites and liquid metal coexist during phase transition like solidification or melting, and the fraction of the solid and liquid phases are determined by the phase diagram of the substance. In our case we assume that concentration variations are negligible and thermophysical and mechanical properties vary continuously with temperature. Although $f_\ell$ is a thermodynamic quantity, in this work it is primarily used as a geometric indicator of the melt-pool boundary via thresholding at 
$f_\ell=0.5$. The melt-pool region is thus defined by $\alpha(\mathbf{x}) \ge 0.5$ and $f_\ell(\mathbf{x}) \ge 0.5$. During the forward passes of the LP-FNO, $f_\ell$ is computed from the inferred $T$ field. It is then aggregated to the supervised loss for the backward pass in order to put extra emphasis on the network to learn the melt-pool boundary temperatures accurately.

\paragraph{Gas phase $\alpha$-masking.} 
\label{sec:alpha_masking}
As the quantities of interest are confined to the metallic domain, and since FLOW-3D WELD\textsuperscript{\textregistered} does not explicitly model the gas phase, a smooth masking procedure is applied to suppress fields in the surrounding gas phase. Specifically, the continuous volume fraction $\alpha(\mathbf{x}) \in [0,1]$ is used to distinguish metal from gas, with $\alpha < 0.5$ indicating regions outside the metal. To preserve differentiability across the metal–gas interface, a smooth interpolation based on the hyperbolic tangent function is defined:
\begin{equation}
g(\mathbf{x}) = \frac{1}{2}\left[\tanh\!\left(k\,(\alpha(\mathbf{x}) - 0.5)\right) + 1\right],
\end{equation}
where $k$ controls the sharpness of the transition (we use $k=20$). The temperature field is then smoothly blended toward a fixed value according to
\begin{equation}
\tilde{T}(\mathbf{x}) = T_{\mathrm{boil}} + g(\mathbf{x})\bigl(T(\mathbf{x}) - T_{\mathrm{boil}}\bigr),
\end{equation}
with $T_{\mathrm{boil}} = 3123~\mathrm{K}$ corresponding to the boiling temperature of Ti–6Al–4V. This operation is applied both after the forward pass of the network during inference and during pre-processing of the ground-truth data, preventing the introduction of spurious sharp gradients at the keyhole interface. During training, the mask is applied after 500 training steps, so that the model has time to learn the general keyhole shape. In addition, the liquid fraction $f_\ell$ and inferred volume fraction $\hat \alpha(\mathbf{x})$  are masked to zero outside the metal, ensuring that they exclusively represent the melt-pool geometry.

\paragraph{Implementation.}
We implemented the FNO, FC-NN, DeepONet, and U-Net models in Python using the PhysicsNeMo \citep{physicsnemo_contributors_nvidia_2023} package from NVIDIA with hyper-parameters described in \autoref{tab:model_arch_hparams} in \autoref{Appendix}, and trained the model on FLOW-3D WELD\textsuperscript{\textregistered} laser welding simulations as described in Section \ref{sec:sim_software}. For training we implement the LION optimizer from \citep{chen_symbolic_2023}. The model hyper-parameters were chosen so that the compared models would be approximately parameter matched with LP-FNO, and further tuned locally to improve accuracy metrics on the validation set.  

\paragraph{K-folds cross validation}
To evaluate the performance of the models, a validation subset was held out from the full dataset and used exclusively for hyperparameter tuning. Owing to the limited number of samples, the remaining data were evaluated using 8-fold cross validation. For each cross-validation split, the model was trained on seven of the eight folds and evaluated on the remaining fold. The reported metrics were then averaged across all eight evaluation folds. To test super-resolution we used the classical train-validation-test set split from \autoref{fig:enthalpy_plots} due to the high computational and storage costs of fine mesh reference multi-physics simulations. To generate the plots in \autoref{fig:error_maps_T}, \ref{fig:fl_segmentation_2x2}, and \ref{fig:alpha_segmentation_2x2}, we used the full dataset with cross-validation, as the errors are reported on a per-simulation basis and any difference across validation folds would be readily apparent. The models converged across all outputs, both in terms of raw loss and in the test set per fold, which is shown in Appendix~\ref{Appendix},  \autoref{fig:fno_training_validation_2x2}.

\paragraph{Metrics for model performance evaluation.}
For evaluating the model performance the mean absolute error (MAE) was used which is defined as follows:
\begin{equation}
\mathrm{MAE} \;=\; \frac{1}{N}\sum_{s=1}^{N} \left(\frac{1}{|D_h|}\sum_{\mathbf{x}\in D_h}\left|e^{(s)}(\mathbf{x})\right|\right),
\label{eq:mae}
\end{equation}
as well as the root mean squared error (RMSE),
\begin{equation}
\mathrm{RMSE} \;=\; \frac{1}{N}\sum_{s=1}^{N} \left(\sqrt{\frac{1}{|D_h|}\sum_{\mathbf{x}\in D_h}\left(e^{(s)}(\mathbf{x})\right)^2}\right).
\label{eq:rmse}
\end{equation}
where $y^{(s)}(\mathbf{x})$ denotes the ground-truth field and $\hat{y}^{(s)}(\mathbf{x})$ the predicted field for sample $s \in \{1,\dots,N\}$ on the discretized domain $D_h$ with $|D_h|$ grid points. The point-wise error is defined as follows:
\begin{equation}
e^{(s)}(\mathbf{x}) = \hat{y}^{(s)}(\mathbf{x}) - y^{(s)}(\mathbf{x}).
\end{equation}
In addition to the MAE and the RMSE, corresponding relative metrics were analyzed. For each output field, a normalization constant was defined as the global mean of the ground-truth values over the evaluation set,
\begin{equation}
\mu_y \;=\; \frac{1}{N\,|D_h|}\sum_{s=1}^{N}\sum_{\mathbf{x}\in D_h} y^{(s)}(\mathbf{x}),
\end{equation}
and relative errors are computed as
\begin{equation}
\mathrm{RelMAE} \;=\; \frac{\mathrm{MAE}}{\mu_y + \varepsilon},
\qquad
\mathrm{RelRMSE} \;=\; \frac{\mathrm{RMSE}}{\mu_y + \varepsilon},
\label{eq:rel_metrics}
\end{equation}
with a small $\varepsilon>0$ used for numerical stability.

For the metal volume fraction $\alpha$ and the liquid fraction $f_\ell$ (which effectively defines the melt-pool boundary), the task can naturally be interpreted as a segmentation problem. Therefore additionally the intersection over union (IoU) was investigated. Given a threshold of $\tau=0.5$, the predicted and ground-truth binary masks for sample $s$ can be defined as:
\begin{equation}
\hat{M}^{(s)}(\mathbf{x}) = \mathbb{I}\!\left[\hat{y}^{(s)}(\mathbf{x}) \ge \tau\right],
\qquad
M^{(s)}(\mathbf{x}) = \mathbb{I}\!\left[y^{(s)}(\mathbf{x}) \ge \tau\right],
\end{equation}
and the IoU as follows:
\begin{equation}
\mathrm{IoU} \;=\; \frac{1}{N}\sum_{s=1}^{N}
\frac{\sum_{\mathbf{x}\in D_h}\hat{M}^{(s)}(\mathbf{x})\, M^{(s)}(\mathbf{x})}
{\sum_{\mathbf{x}\in D_h}\mathbb{I}\!\left[\hat{M}^{(s)}(\mathbf{x}) + M^{(s)}(\mathbf{x}) \ge 1\right]}.
\label{eq:iou}
\end{equation}
Relative metrics were not computed for $\alpha$ and $f_\ell$ since these fields are already dimensionless and consist largely of $0$ valued points; thus, they are evaluated primarily through absolute errors and IoU.

\section{Results and Discussion}

\begin{figure}[h]
    \centering
    \includegraphics[width=\linewidth]{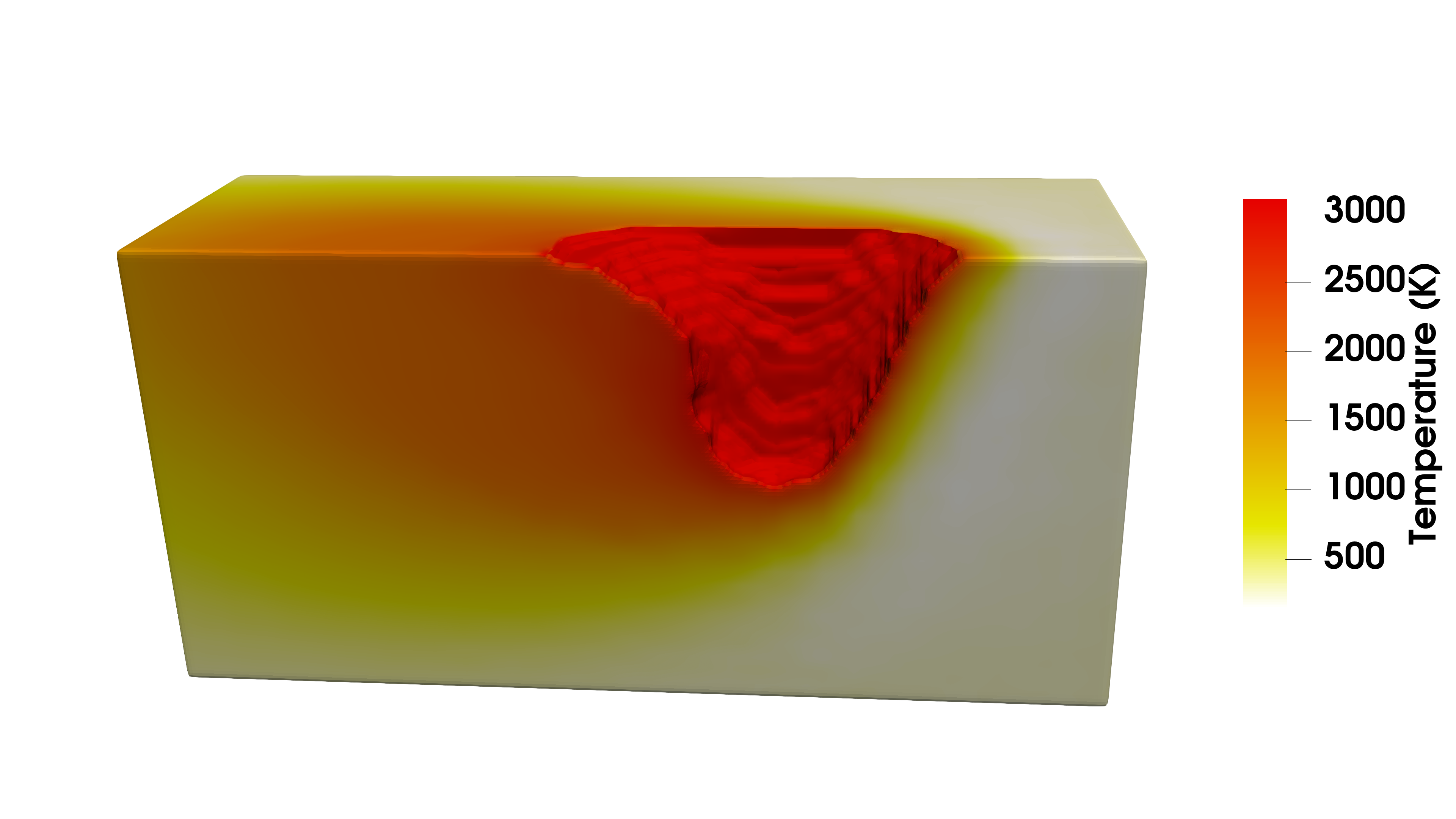}
    \caption{LP-FNO 3D inference result for keyhole regime simulation with $P=\SI{150}{\watt}$ and $V_{\mathrm{scan}}=\SI{0.54}{\meter\per\second}$. Domain cut in half for visual clarity. Visualized using ParaView \citep{ahrens_paraview_2005}.}
    \label{fig:fno_inference_3D}
\end{figure}

Once the LP-FNO has been trained, we can generate the temperature field and melt-pool from any set of process parameters within the training range in a fraction of a second for any desired mesh resolution. For example, if we input the power $P=\SI{150}{\watt}$ and laser scanning speed $V_{scan}=\SI{0.54}{\meter\per\second}$, we get the temperature field and melt-pool shown in \autoref{fig:fno_inference_3D} in around \SI{8}{\milli \s} for \SI{10}{\micro \meter} resolution and in around \SI{88}{\milli \s} for \SI{5}{\micro \meter} resolution. This is more than ten thousand times faster than the FLOW-3D\textsuperscript{\textregistered} simulation equivalent at \SI{10}{\micro \meter} resolution and around one hundred thousand times faster than FLOW-3D\textsuperscript{\textregistered} at \SI{5}{\micro \meter} resolution. 
\paragraph{Runtime comparison.}
The inference and training runtimes reported in \autoref{tab:runtime_comparison} show that LP-FNO is significantly faster than the tested neural surrogate architectures with comparable trainable parameter counts. Although the FC-NN and U-Net architectures contain approximately three times more trainable parameters than LP-FNO, their runtimes are substantially more than three times larger. The two DeepONet architectures, by contrast, have almost the same number of trainable parameters as LP-FNO, making the runtime comparison particularly direct. The corresponding architectures and hyperparameters are summarized in \autoref{tab:model_arch_hparams}. The difference is most pronounced for inference on the super-resolved \SI{5}{\micro\meter} grid. Pointwise architectures, such as FC-NN and DeepONet, must evaluate individual spatial locations, whereas grid-based models process the full computational domain in a single forward pass, leading to substantially lower inference times on large domains. Moreover, as exemplified by its poor super-resolution accuracy in \autoref{tab:metrics_superresolved_comparison}, the U-Net architecture is not intrinsically resolution-invariant. Its \SI{5}{\micro\meter} runtime should therefore not be interpreted as meaningful super-resolved inference in the same sense as LP-FNO. Overall, among the tested surrogate models capable of super-resolved inference, LP-FNO is approximately two orders of magnitude faster on the \SI{5}{\micro\meter} grid.

\begin{table*}[h!]
\centering
\caption{Comparison of computational runtimes for high-fidelity FLOW-3D\textsuperscript{\textregistered} simulations and neural surrogate models. FLOW-3D\textsuperscript{\textregistered} runtimes correspond to transient simulations at power \SI{50}{\watt}, scan speed \SI{0.260}{\meter\per\second}, and scan distance \SI{0.6}{mm}. Surrogate runtimes report one-time training cost, where available, and per-sample inference time. Inference timings were measured on an Ubuntu 24.04.3 LTS workstation with an Intel Core i9-10900X CPU, 125 GiB RAM, and dual NVIDIA TITAN RTX GPUs, although only one GPU was used at a time. FLOW-3D\textsuperscript{\textregistered} simulations were run on an AMD Ryzen Threadripper 7960X using 16 cores. The fastest times are marked in bold.}
\label{tab:runtime_comparison}
\scriptsize
\setlength{\tabcolsep}{5pt}
\renewcommand{\arraystretch}{1.1}
\begin{tabular}{llccc}
\toprule
\textbf{Method} 
& \textbf{Data layout} 
& \textbf{Training time} 
& \textbf{\SI{10}{\micro\meter} mesh} 
& \textbf{\SI{5}{\micro\meter} mesh} \\
\midrule
FLOW-3D\textsuperscript{\textregistered} simulation
& -- 
& -- 
& \SI{6}{\minute}~\SI{7}{\second} 
& \SI{1}{\hour}~\SI{6}{\minute} \\

FC-NN inference
& pointwise 
&  \SI{2}{\hour}~\SI{52}{\minute}
& \SI{2.21}{\second} 
& \SI{32.5}{\second} \\

DPN-FCT inference
& pointwise 
& \SI{1}{\hour}~\SI{5}{\minute} 
& \SI{803}{\milli\second} 
& \SI{15.6}{\second} \\

DPN-FourierT inference
& pointwise 
& \SI{1}{\hour}~\SI{42}{\minute}
& \SI{385}{\milli\second} 
& \SI{7.42}{\second} \\

U-Net inference
& grid 
& \SI{52}{\minute}~\SI{13}{\second}
& \SI{51.1}{\milli\second} 
& \SI{861}{\milli\second} \\

LP-FNO inference
& grid 
& \textbf{\SI{8}{\minute}~\SI{20}{\second}} 
& \textbf{\SI{7.70}{\milli\second}} 
& \textbf{\SI{87.9}{\milli\second}} \\
\bottomrule
\end{tabular}
\end{table*}

\begin{figure}[h]
    \centering
    \begin{subfigure}[t]{0.49\textwidth}
        \centering
        \includegraphics[width=\textwidth]{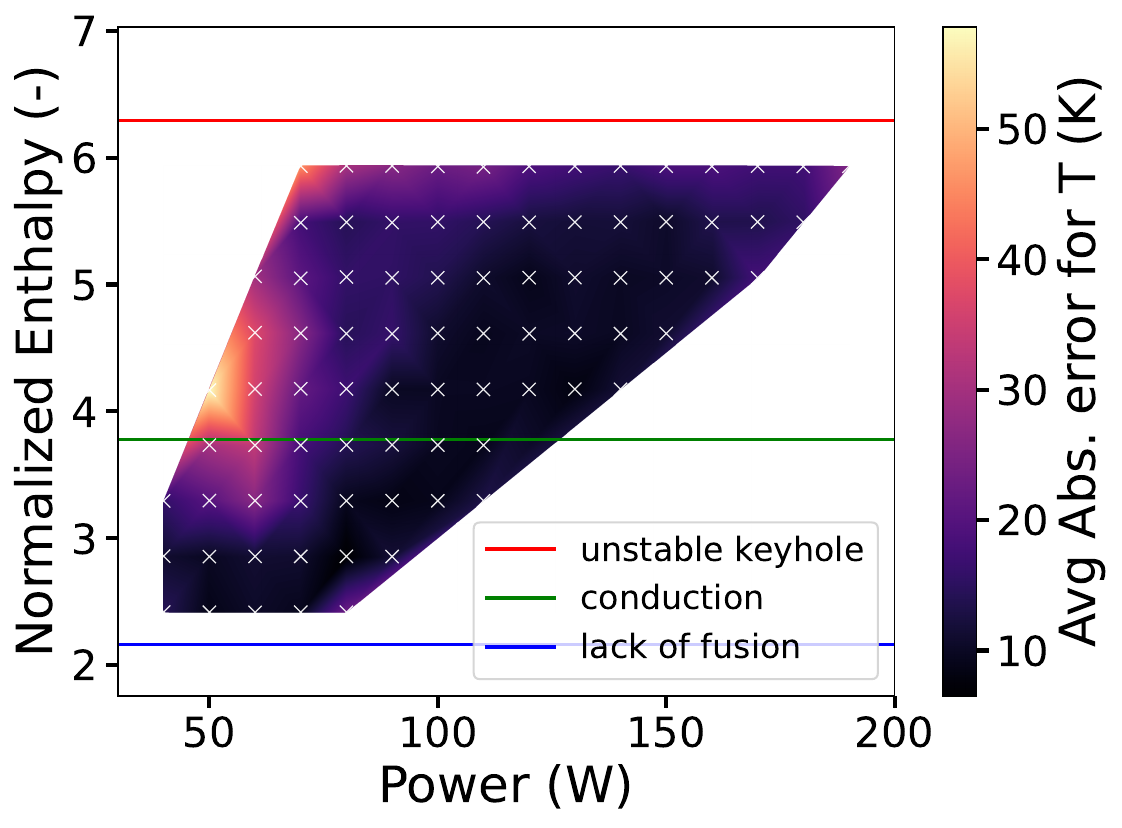}
        \caption{Average absolute temperature error (Kelvin) as a function of laser power $P$ and normalized enthalpy $H^\ast$}
        \label{fig:error_HP}
    \end{subfigure}
    \hfill
    \begin{subfigure}[t]{0.49\textwidth}
        \centering
        \includegraphics[width=\textwidth]{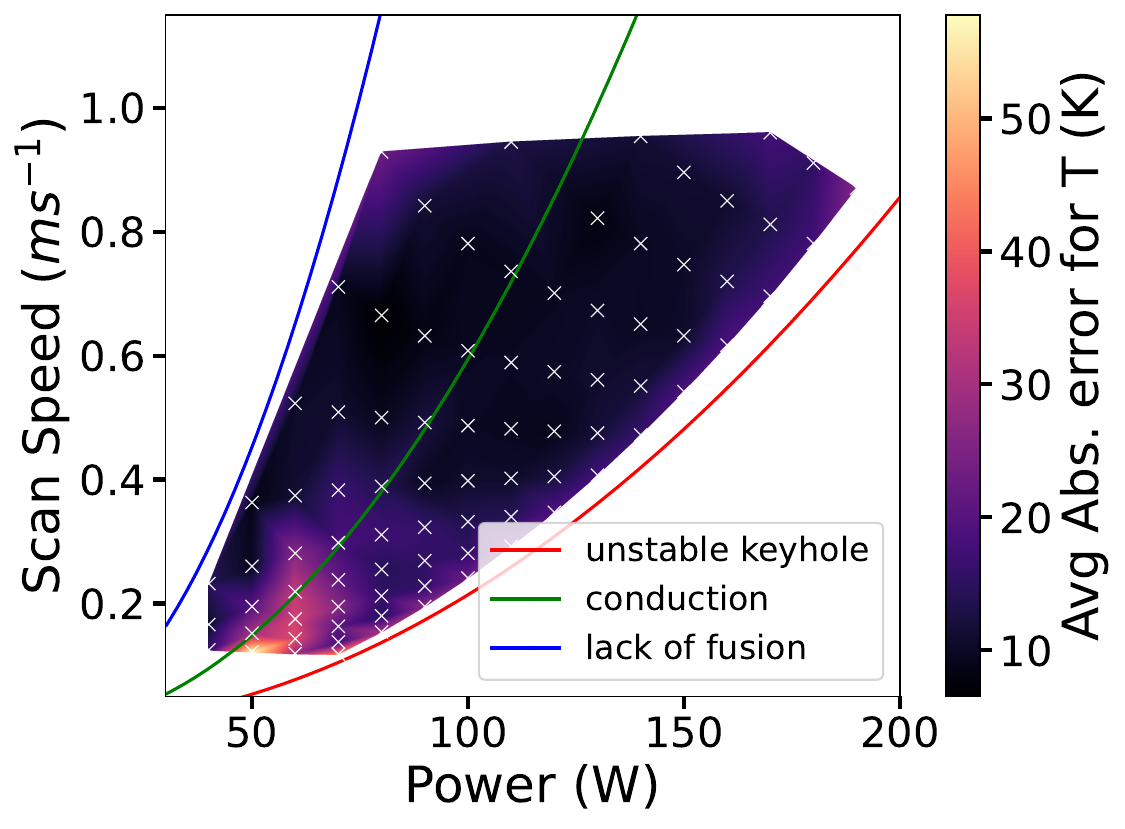}
        \caption{Average absolute temperature error (Kelvin) as a function of laser power $P$ and scan speed $V_{\text{scan}}$}
        \label{fig:error_PV}
    \end{subfigure}

    \caption{Model error maps over the process parameter space. The color scale indicates the average absolute error for the predicted temperature field. Solid lines denote approximate regime indicators corresponding to lack of fusion (blue), conduction (green), and unstable keyhole (red). White markers indicate simulation samples used for training and evaluation with k-folds cross validation}
    \label{fig:error_maps_T}
\end{figure}

\begin{figure}[h]
    \centering
    \begin{subfigure}[b]{0.48\linewidth}
        \centering
        \includegraphics[width=\linewidth]{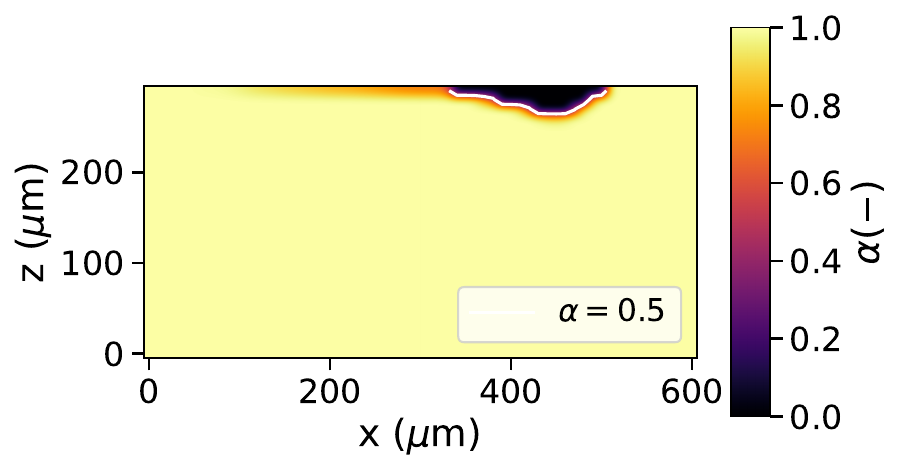}
        \caption{Ground truth}
        \label{fig:outlier_alpha_gt}
    \end{subfigure}
    \hfill
    \begin{subfigure}[b]{0.48\linewidth}
        \centering
        \includegraphics[width=\linewidth]{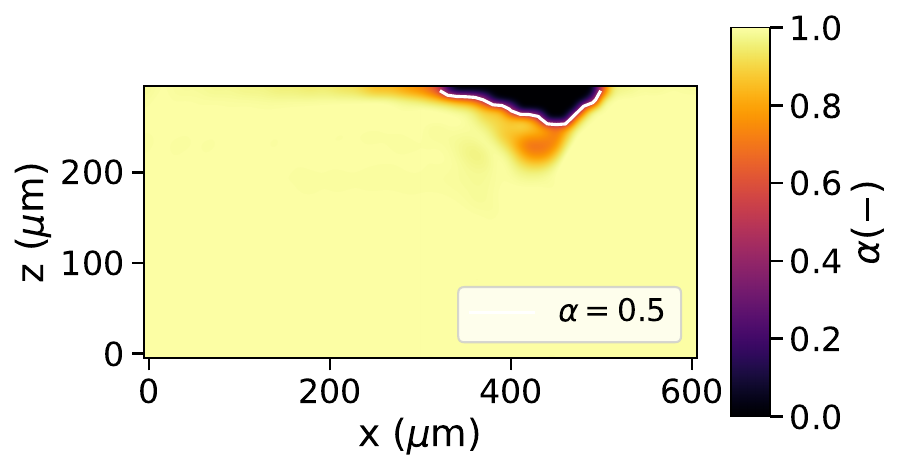}
        \caption{LP-FNO prediction}
        \label{fig:outlier_alpha_pred}
    \end{subfigure}

    \caption{Gas--metal interface ($\alpha$) ground truth and FNO inference for an outlier case at $P=\SI{50}{\watt}$ and $V_{\mathrm{scan}}=\SI{0.122}{\meter\per\second}$, where the model exhibits its largest deviation. This operating point lies close to the transition between conduction and keyhole regimes, where the model exhibits increased uncertainty. The model begins to smear a small keyhole-like feature in the $\alpha$ field, which also leads to smearing in the temperature field because the model is uncertain about the local regime. A few additional training data points in this region would likely greatly reduce this error}
    \label{fig:outlier_alpha}
\end{figure}

\begin{figure}[h]
    \centering
    \begin{subfigure}[b]{0.48\textwidth}
        \centering
        \includegraphics[width=\textwidth]{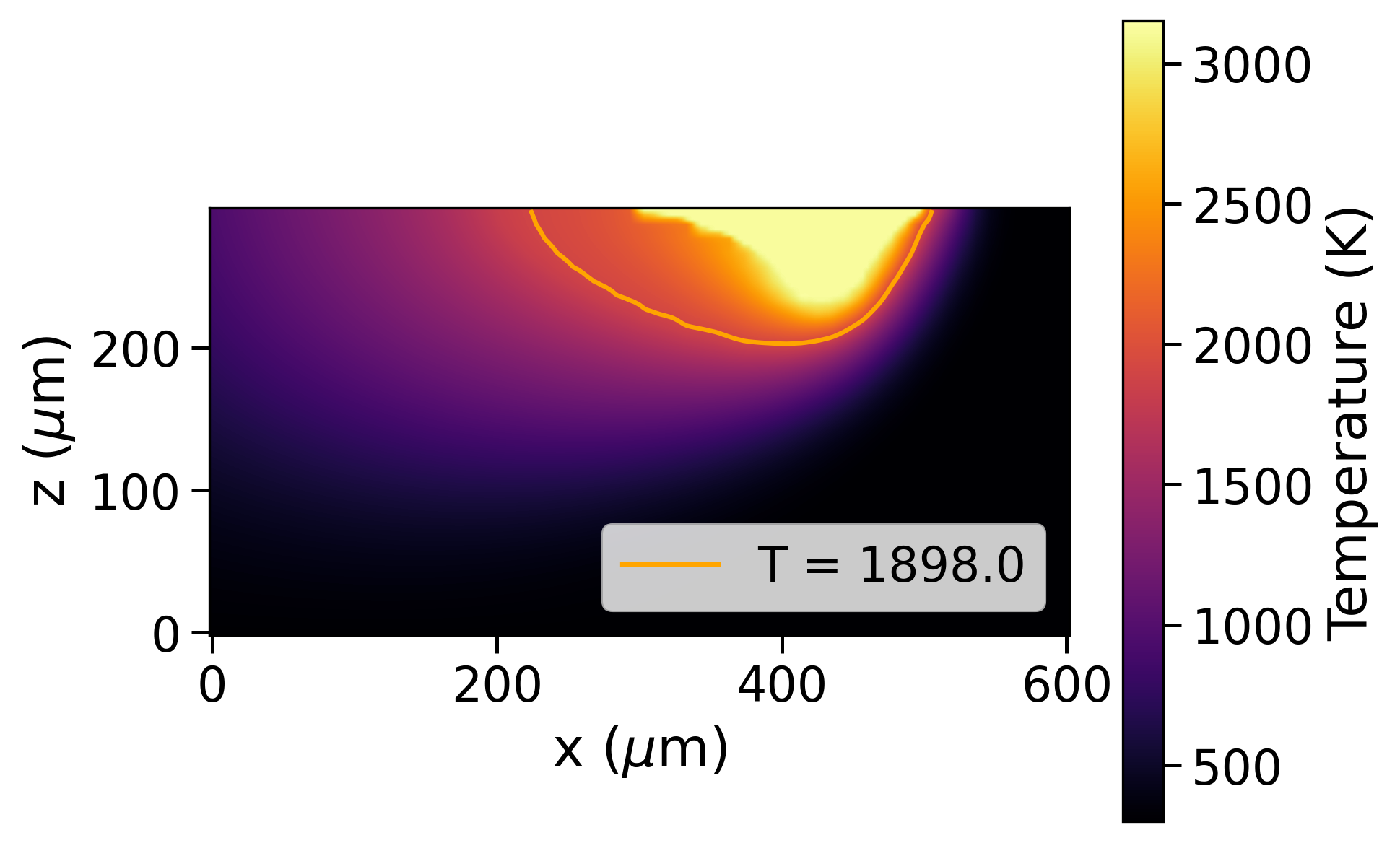}
        \caption{FLOW-3D\textsuperscript{\textregistered}, coarse mesh (\SI{10}{\micro\meter})}
        \label{fig:conduction_flow3d_coarse}
    \end{subfigure}
    \hfill
    \begin{subfigure}[b]{0.48\textwidth}
        \centering
        \includegraphics[width=\textwidth]{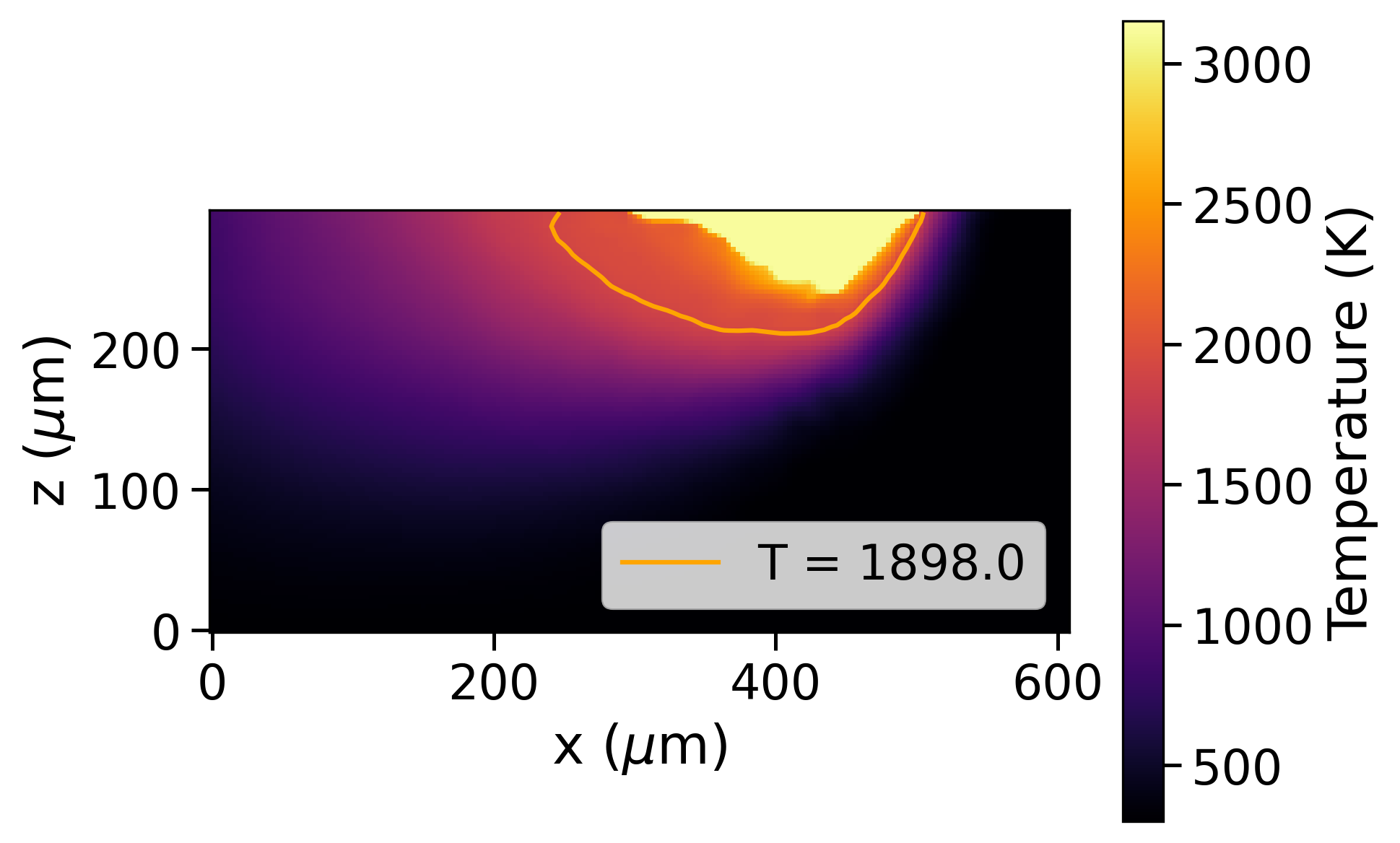}
        \caption{FNO prediction, coarse mesh (\SI{10}{\micro\meter})}
        \label{fig:conduction_fno_coarse}
    \end{subfigure}

    \vspace{2mm}

    \begin{subfigure}[b]{0.48\textwidth}
        \centering
        \includegraphics[width=\textwidth]{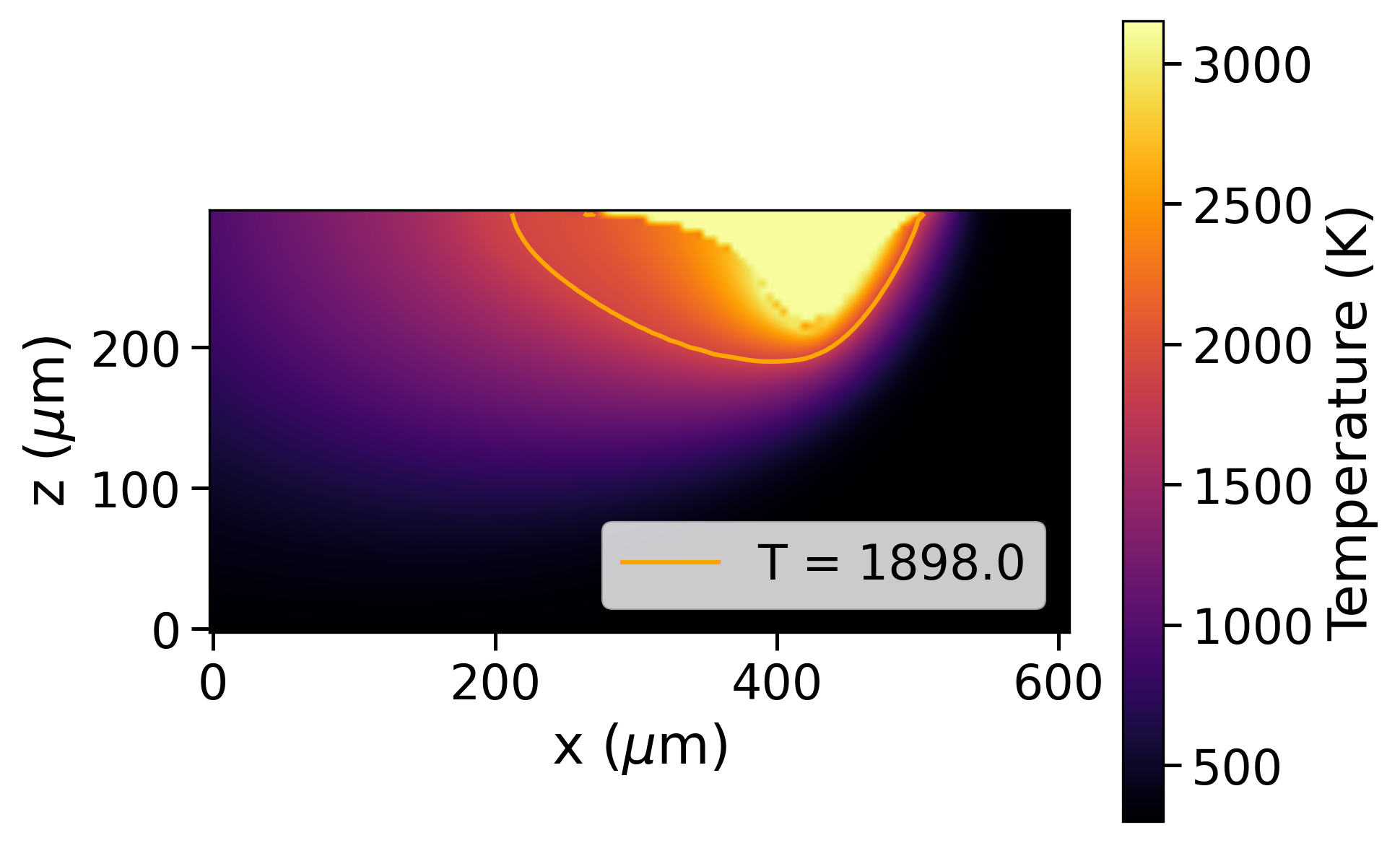}
        \caption{FLOW-3D\textsuperscript{\textregistered}, fine mesh (5~$\mu$m)}
        \label{fig:conduction_flow3d_fine}
    \end{subfigure}
    \hfill
    \begin{subfigure}[b]{0.48\textwidth}
        \centering
        \includegraphics[width=\textwidth]{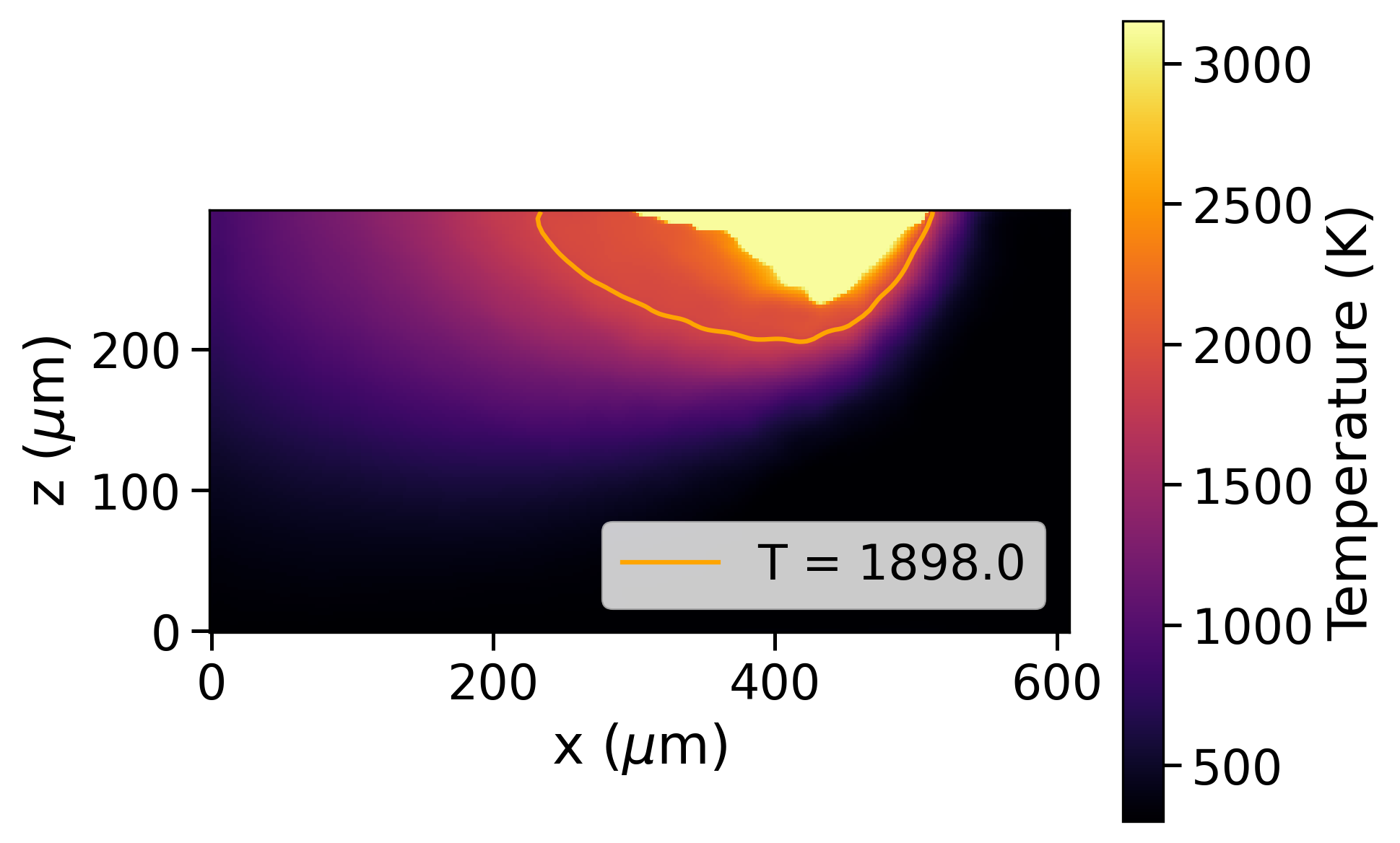}
        \caption{FNO super-resolved prediction (5~$\mu$m)}
        \label{fig:conduction_fno_fine}
    \end{subfigure}

    \caption{Conduction-regime melt-pool mid-point cross-section comparison for a test case with $P=70.0$ and $V_{\mathrm{scan}}=0.298~\mathrm{m\,s^{-1}}$.  
    In contrast to the keyhole regime, the coarse and fine FLOW-3D\textsuperscript{\textregistered} solutions are visually similar, indicating that mesh convergence is effectively achieved at \SI{10}{\micro\meter} resolution in the conduction regime}
    \label{fig:conduction_coarse_fine_comparison}
\end{figure}

\begin{figure}[h]
    \centering
    \begin{subfigure}[b]{0.48\textwidth}
        \centering
        \includegraphics[width=\textwidth]{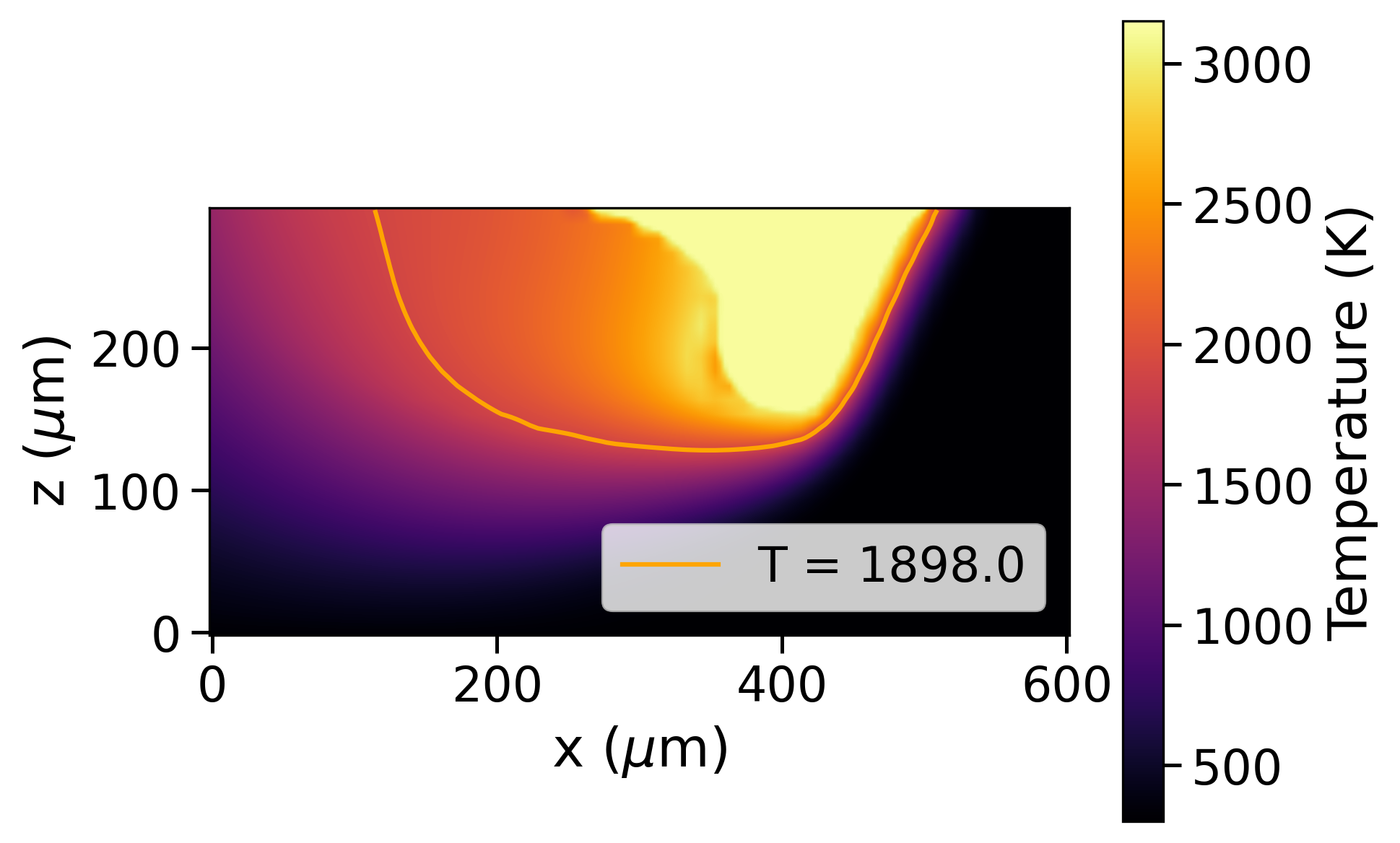}
        \caption{FLOW-3D\textsuperscript{\textregistered}, coarse mesh (\SI{10}{\micro\meter})}
        \label{fig:keyhole_flow3d_coarse}
    \end{subfigure}
    \hfill
    \begin{subfigure}[b]{0.48\textwidth}
        \centering
        \includegraphics[width=\textwidth]{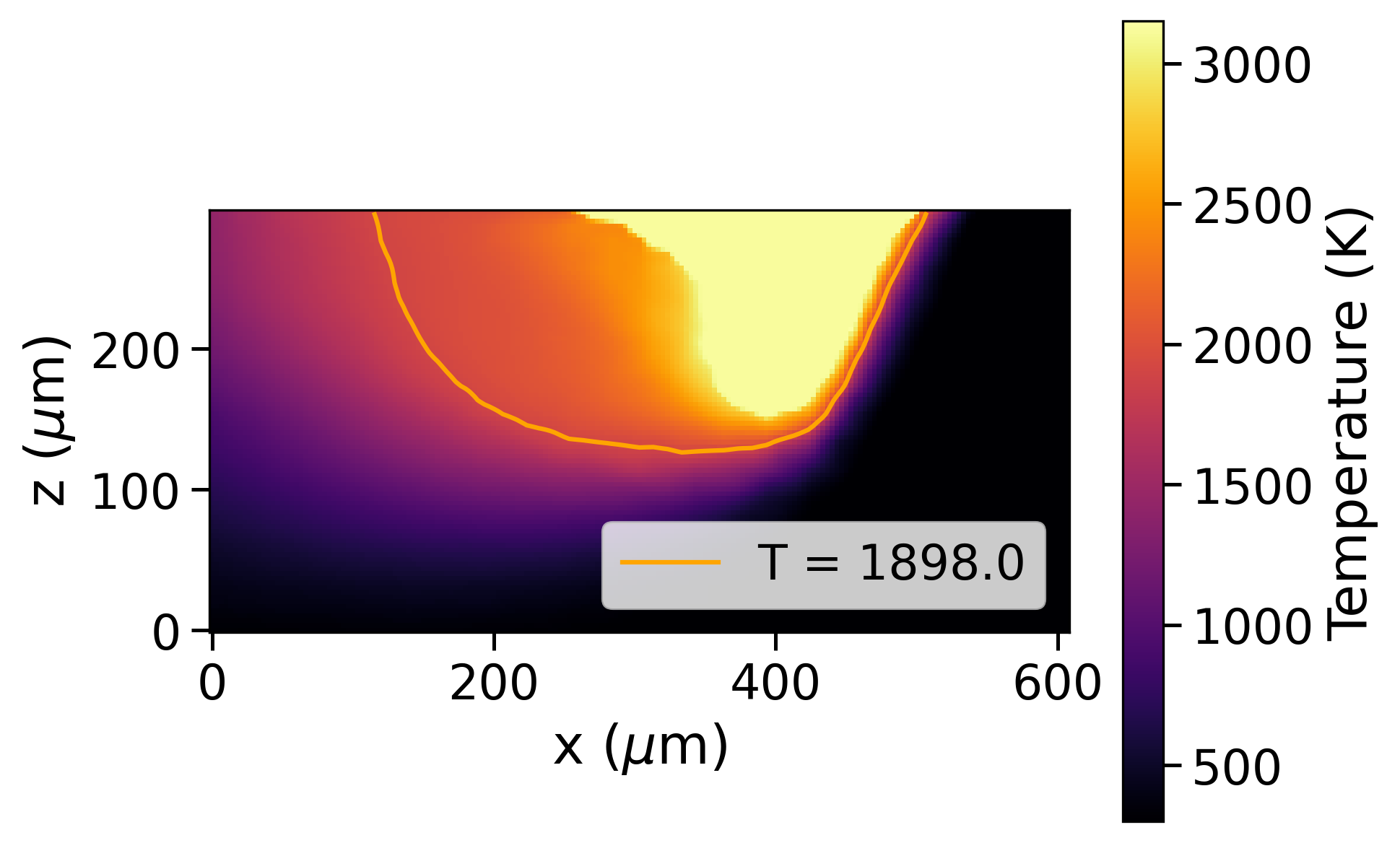}
        \caption{FNO prediction, coarse mesh (\SI{10}{\micro\meter})}
        \label{fig:keyhole_fno_coarse}
    \end{subfigure}

    \vspace{2mm}

    \begin{subfigure}[b]{0.48\textwidth}
        \centering
        \includegraphics[width=\textwidth]{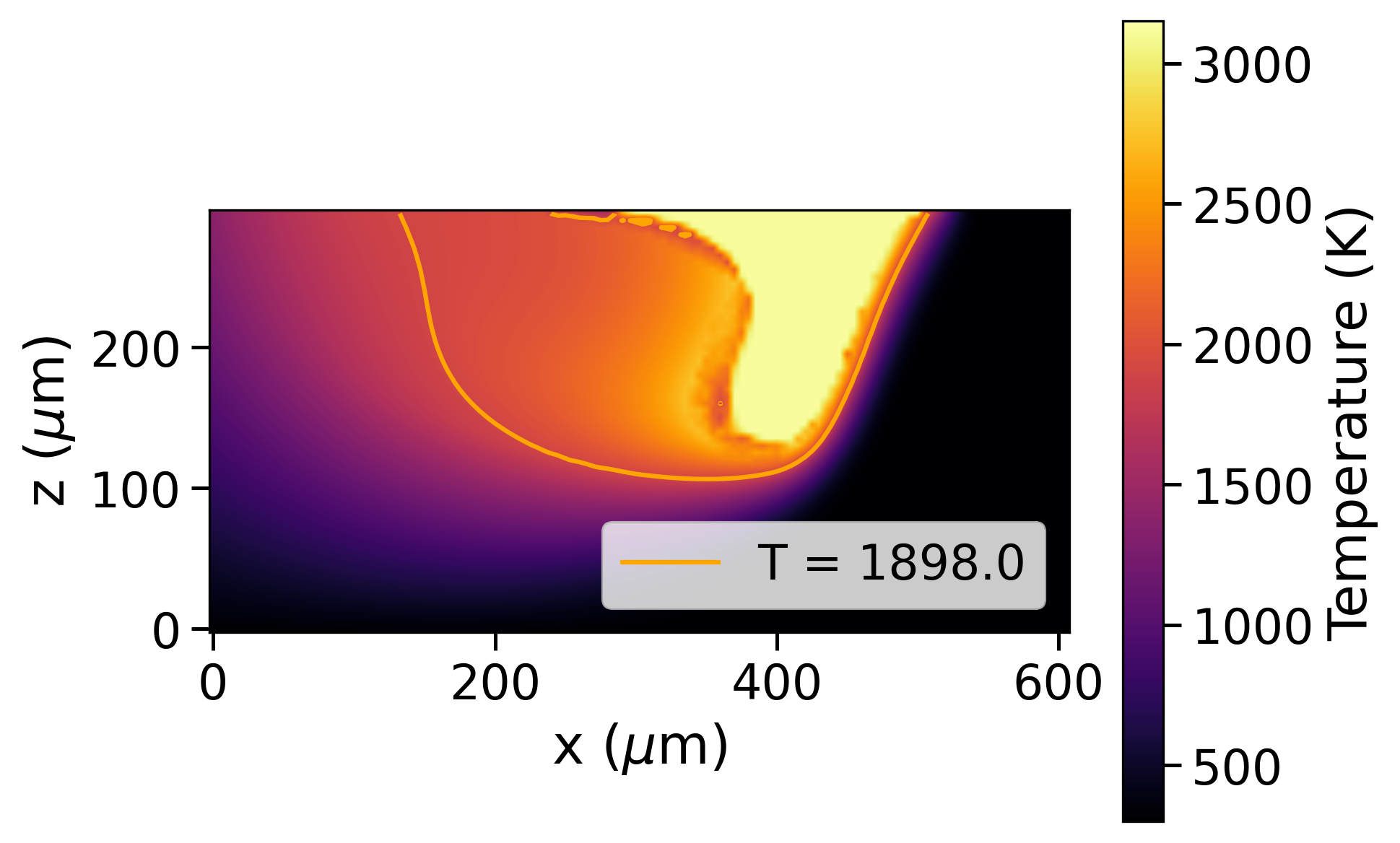}
        \caption{FLOW-3D\textsuperscript{\textregistered}, fine mesh (\SI{5}{\micro\meter})}
        \label{fig:keyhole_flow3d_fine}
    \end{subfigure}
    \hfill
    \begin{subfigure}[b]{0.48\textwidth}
        \centering
        \includegraphics[width=\textwidth]{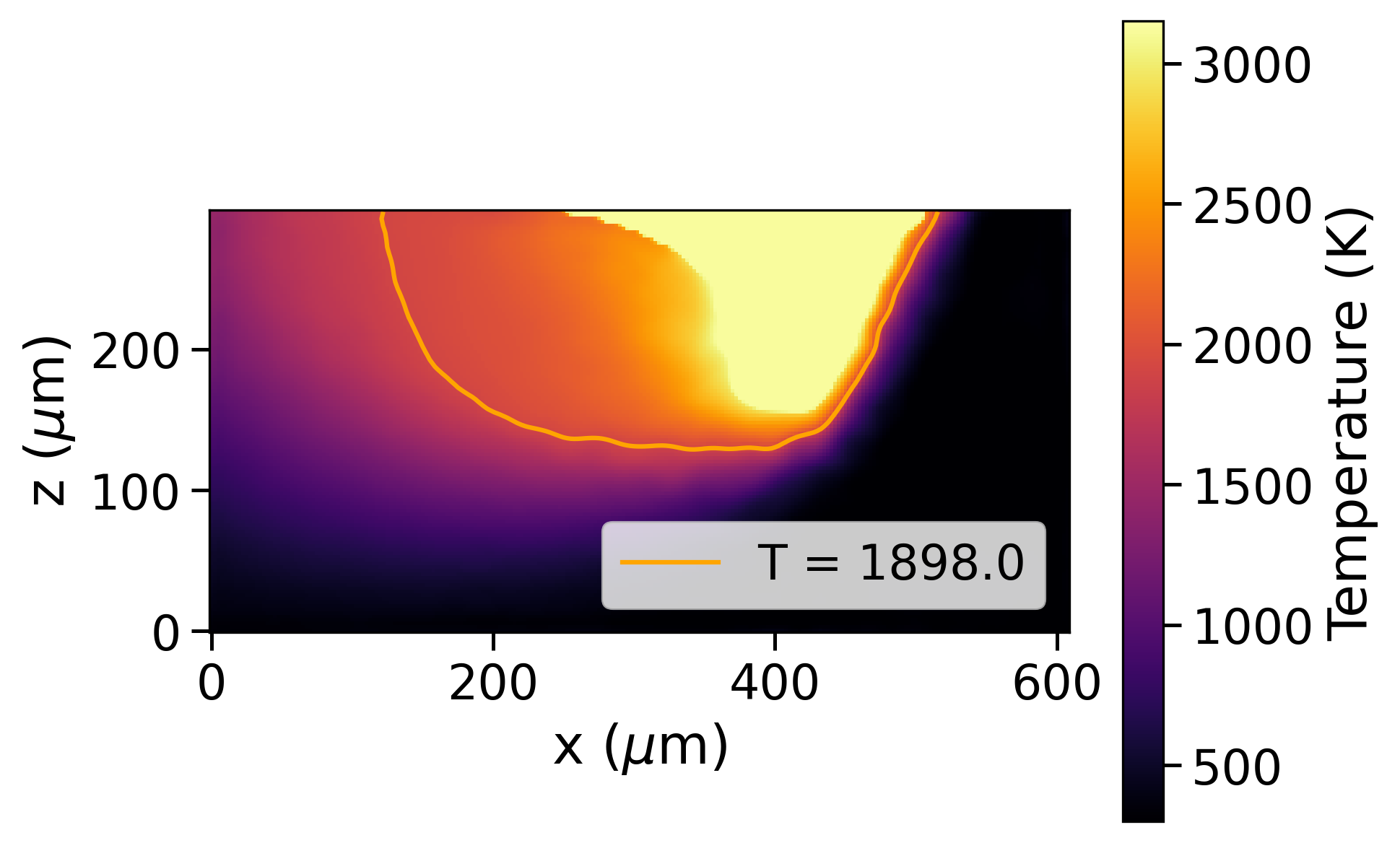}
        \caption{FNO super-resolved prediction (\SI{5}{\micro\meter})}
        \label{fig:keyhole_fno_fine}
    \end{subfigure}

    \caption{Keyhole-regime melt-pool mid-point cross-section comparison for a test case with $P=\SI{150.0}{\watt}$ and $V_{\mathrm{scan}}=\SI{0.54}{\meter\per\second}$.  
    The fine-resolution FNO result is obtained by super-resolving a model trained exclusively on coarse (\SI{10}{\micro\meter}) data. 
    The visible discrepancy between coarse and fine FLOW-3D solutions indicates that mesh convergence has not yet been achieved at \SI{10}{\micro\meter} resolution in the keyhole regime. This shows that while useful, FNO super-resolution can only be as good as its training data fidelity}
    \label{fig:keyhole_coarse_fine_comparison}
\end{figure}

\paragraph{Metrics across different parameter combinations for \SI{10}{\micro\meter} mesh.}
To evaluate how well our model performs over different parameter combinations, we first consider our metrics as a heat map over the entire parameter domain using K-folds to identify areas in the map where the model performs particularly well or poorly. We see in \autoref{fig:error_maps_T} that there appears to be one outlier point at $P=\SI{50}{\watt}$, $V_{scan}=\SI{0.122}{\meter\per\second}$, which lies at edge of our data in start of conduction-keyhole transition. Looking more closely at the inference of $\alpha$ on that point in \autoref{fig:outlier_alpha}, we see that as this is a transition point, the model is unsure as to whether this is a keyhole point of not, leading to some smearing of the interface. Within this region of parameter space, the keyhole geometry is very sensitive to process parameters. Combined with the lack of data in the region, this makes accurate predictions harder for ML models, leading to higher errors as seen in \autoref{fig:error_maps_T}. 
Other than at this point, the model performs quite strongly with most points on the map having an average temperature error of \SI{10}{\kelvin}, with the error rising slightly to around \SI{25}{\kelvin} at the high enthalpy (low speed, high power) keyhole region of the map, as anticipated due to the increased complexity of these regions and the lack of training data on the edges of the map. 
Indeed, if we were to extrapolate the model to normalized enthalpy levels higher than around 6, we would move beyond the onset of unstable keyhole behavior which would be unsuitable for our current quasi-steady formulation. In this high enthalpy region, the time-averaged representation ceases to be a faithful surrogate for the instantaneous melt-pool state. Critically, porosity nucleation, spatter ejection, and the abrupt loss of keyhole stability are inherently non-equilibrium phenomena whose signatures are destroyed by the averaging operation, and a quasi-steady surrogate cannot, by construction, predict them. 

The segmentation of the melt-pool ($f_\ell$) in \autoref{fig:fl_segmentation_2x2} and the metal-gas interface $\alpha$ in \autoref{fig:alpha_segmentation_2x2} in \autoref{Appendix}, demonstrate a similar pattern. The same outlier is present and the high enthalpy, keyhole edge of the map has higher error, except for the melt-pool IoU error. This is expected since the melt-pool is larger in the high enthalpy regime, which shrinks small segmentation errors in comparison to the same error in a small melt-pool. Similar reasoning explains why IoU errors in \autoref{fig:fl_iou_ph} and \ref{fig:fl_iou_pv} have comparatively large magnitudes, especially in the low power and low enthalpy regimes, since melt-pools in those regimes can be very small, an error of even one voxel makes up a few percents of the total melt-pool and is thus amplified.

\begin{table*}[h!]
\centering
\caption{Comparison of model error metrics reported as fold means across 8 folds, excluding the validation set used for hyperparameter tuning. Absolute temperature errors are reported in kelvin after multiplying the normalized values by $T_{ref}=3000$, and relative temperature errors are reported in percent. Relative metrics are not reported for $\alpha$ and $f_\ell$ because they are already dimensionless. Best values are highlighted in bold; lower is better except for IoU mean.}
\label{tab:model_metrics_comparison}
\scriptsize
\setlength{\tabcolsep}{4pt}
\renewcommand{\arraystretch}{1.1}
\resizebox{\textwidth}{!}{%
\begin{tabular}{lccccc}
\toprule
\textbf{Metric} & \textbf{LP-FNO} & \textbf{DPN-FCT} & \textbf{DPN-FourierT} & \textbf{FC-NN} & \textbf{U-Net} \\
\midrule
$T$ abs. mean err. [K] 
& \textbf{15.7} & 30.1 & 37.3 & 19.2 & 27.1 \\

$T$ RMSE [K] 
& \textbf{33.5} & 57.0 & 64.6 & 34.8 & 44.6 \\

$T$ rel. mean  err. [\%] 
& \textbf{2.49} & 4.78 & 5.91 & 3.05 & 4.30 \\

$T$ rel. RMSE [\%] 
& \textbf{5.31} & 9.05 & 10.26 & 5.52 & 7.08 \\

\midrule
$\alpha$ abs. mean err.
& \textbf{0.0024} & 0.0072 & 0.0102 & 0.0028 & 0.0097 \\

$\alpha$ RMSE 
& 0.0163 & 0.0233 & 0.0254 & \textbf{0.0158} & 0.0257 \\

$\alpha$ IoU mean 
& \textbf{0.9992} & 0.9989 & 0.9989 & 0.9990 & 0.9990 \\

\midrule
$f_\ell$ abs. mean err.
& \textbf{0.0020} & 0.0110 & 0.0132 & 0.0029 & 0.0028 \\

$f_\ell$ RMSE 
& \textbf{0.0327} & 0.0985 & 0.1087 & 0.0375 & 0.0397 \\

$f_\ell$ IoU mean 
& \textbf{0.9216} & 0.7016 & 0.6629 & 0.9044 & 0.8966 \\

\bottomrule
\end{tabular}%
}
\end{table*}

\paragraph{Average metrics for \SI{10}{\micro\meter} mesh.}
More generally, \autoref{tab:model_metrics_comparison} shows that LP-FNO performs well across all reported metrics. For the temperature field, LP-FNO achieves an average absolute error of \SI{15.7}{\kelvin}, corresponding to approximately $2.5\%$, and an RMSE of \SI{33.5}{\kelvin}, corresponding to $5.31\%$. For the metal--gas interface $\alpha$, the average absolute error is $0.0024$, with an RMSE of $0.0163$ and a near-perfect IoU of $0.9992$. For the full melt-pool segmentation, the average absolute error is $0.0020$, with an RMSE of $0.0327$ and an IoU of $0.922$. The discrepancy between the $\alpha$ and $f_\ell$ IoU values is expected, since the melt-pool region denoted by $f_\ell$ is much smaller than the full metal region denoted by $\alpha$. Consequently, small absolute segmentation errors represent a larger fraction of the melt-pool volume and therefore have a stronger effect on the $f_\ell$ IoU. The moderate increase from MAE to RMSE indicates that localized regions of larger error are present, but that these errors remain limited and confined to specific regions on the boundaries of the process space, as confirmed by \autoref{fig:error_maps_T}, \autoref{fig:alpha_segmentation_2x2}, and \autoref{fig:fl_segmentation_2x2}.

\paragraph{Accuracy comparison of different model architectures.}
The accuracy comparison with the different architectures in \autoref{tab:model_metrics_comparison} further indicates that LP-FNO provides the strongest overall performance for the present task. It achieves the lowest temperature errors across all reported temperature metrics and the best melt-pool segmentation performance, with the highest $f_\ell$ IoU and lowest $f_\ell$ RMSE. The FC-NN model is also surprisingly competitive despite its simpler architecture, particularly for temperature prediction and $\alpha$ reconstruction, suggesting that a substantial part of the mapping can be learned from local coordinate and process-parameter information. However, LP-FNO remains more consistent across the different output fields, especially for the melt-pool boundary. 
While FC-NN achieves competitive accuracy for several temperature and interface metrics, it is substantially slower at inference, especially on the finer grid. This indicates that the main advantage of LP-FNO is not only its accuracy, but its combined accuracy, computational efficiency, and resolution-invariant evaluation. These results should not, however, be interpreted as a definitive statement that FNOs are universally superior to the tested models. The accuracy differences between all the models are relatively small, and further architecture design or more extensive hyperparameter tuning could plausibly reduce the gap. Nevertheless, under the training protocol, parameter ranges, and model configurations considered in this study, the FNO-based architecture provides the most favorable accuracy--efficiency trade-off.

\paragraph{Super Resolution.}
One of the strongest features of FNO is that, as described in Section \ref{sec:super-resolution}, it is resolution independent. We can train FNO on coarse simulation data, which is much faster to obtain (time and data size grow cubically with mesh refinement), and perform inference of the model on a much finer mesh with runtime that remains within the same practical real-time range. We evaluate our model trained on simulations with mesh size \SI{10}{\micro\meter} on a representative sample with mesh size \SI{5}{\micro\meter}, covering all physical regimes from conduction to keyhole depicted by the test set points in \autoref{fig:enthalpy_plots}. We do not run K-folds on the entire grid here due to the high computational and storage costs of fine mesh reference multi-physics simulations. This super-resolution evaluation should therefore be interpreted as a dedicated test-set study rather than a full cross-validated assessment.

We can see the comparison between the coarse and fine mesh simulation and inference results for a conduction regime parameter set in \autoref{fig:conduction_coarse_fine_comparison} and for a keyhole regime set in \autoref{fig:keyhole_coarse_fine_comparison}. In both cases, the super-resolved FNO inference is essentially a finer representation of the coarse inference result. This is expected, and as such, the result is as accurate as the training set was accurate. In other words, the FNO enables numerical super-resolution, but it cannot recover physical features that are absent from the coarse-mesh training data. Hence, we have good agreement in the conduction mode simulation, since the corresponding coarse and fine FLOW-3D\textsuperscript{\textregistered} simulations are close to each other. However, for the keyhole mode, we see that the fine FLOW-3D\textsuperscript{\textregistered} simulation is slightly more complex than the coarse counterpart, and hence, the super-resolved FNO inference result cannot fully capture the complexity of the fine FLOW-3D\textsuperscript{\textregistered} reference simulation. The discrepancy in the keyhole regime is therefore primarily attributed to the lack of mesh convergence of the \SI{10}{\micro\meter} training data in this regime, rather than to the neural operator alone.

The super-resolution metrics in \autoref{tab:metrics_superresolved_comparison} show that the errors in the super-resolved case are roughly double those of the non-super-resolved case in \autoref{tab:model_metrics_comparison}. As discussed, the discrepancy primarily comes from the lack of mesh convergence for the \SI{10}{\micro\meter} mesh size training data in the keyhole regime. Compared with the tested architectures, LP-FNO provides the strongest overall super-resolution performance, achieving the lowest temperature mean error, the lowest $f_\ell$ error, and the highest $f_\ell$ IoU. The FC-NN model remains competitive for temperature prediction and even obtains the lowest temperature RMSE, but LP-FNO is more consistent across the temperature, metal--gas interface, and melt-pool outputs. Moreover, the U-Net architecture is demonstrated not to be intrinsically resolution-invariant, as reflected by its poor super-resolution accuracy in \autoref{tab:metrics_superresolved_comparison}. One possible strategy to improve physical fidelity in all regimes, while keeping computational training cost reasonable, would be to employ a coarse-to-fine training strategy, where FNO is first trained on a large set of coarse-mesh simulations, then progressively fine-tuned on finer meshed data for regimes where training data has not yet reached mesh convergence. Such a multi-fidelity strategy could preserve the low cost of coarse simulations while improving the fidelity of the surrogate in regimes where fine-mesh effects are physically important.

\begin{table*}[h!]
\centering
\caption{Comparison of model error metrics on the super-resolution test data evaluated on a \SI{5}{\micro\meter} mesh. Absolute temperature errors are reported in Kelvin, and relative temperature errors are reported in percent. Best values are highlighted in bold; lower is better except for IoU mean. For U-Net, the \SI{5}{\micro\meter} evaluation is included only to illustrate the failure of direct resolution transfer for a grid-bound architecture; it should not be interpreted as valid super-resolved inference.}
\label{tab:metrics_superresolved_comparison}
\scriptsize
\setlength{\tabcolsep}{4pt}
\renewcommand{\arraystretch}{1.1}
\resizebox{\textwidth}{!}{%
\begin{tabular}{lccccc}
\toprule
\textbf{Metric} & \textbf{LP-FNO} & \textbf{DPN-FCT} & \textbf{DPN-FourierT} & \textbf{FC-NN} & \textbf{U-Net} \\
\midrule
$T$ abs. mean err. [K] 
& \textbf{27.7} & 32.7 & 83.3 & 28.2 & \num{2.75e4} \\

$T$ RMSE [K] 
& 65.5 & 67.5 & 196 & \textbf{59.0} & \num{2.67e5} \\

$T$ rel. mean err. [\%] 
& \textbf{4.58} & 5.40 & 13.8 & 4.67 & \num{4.55e3} \\

$T$ rel. RMSE [\%] 
& 10.8 & 11.2 & 32.5 & \textbf{9.75} & \num{4.42e4} \\

\midrule
$\alpha$ abs. mean err. [-]
& \textbf{0.00361} & 0.00630 & 0.0109 & 0.00525 & 18.9 \\

$\alpha$ RMSE [-]
& 0.0357 & \textbf{0.0332} & 0.0725 & 0.0354 & 196 \\

$\alpha$ IoU mean [-]
& \textbf{0.998} & \textbf{0.998} & 0.994 & \textbf{0.998} & 0.991 \\

\midrule
$f_\ell$ abs. mean err. [-]
& \textbf{0.00750} & 0.0134 & 0.0237 & 0.00797 & 0.0321 \\

$f_\ell$ RMSE [-]
& \textbf{0.0764} & 0.110 & 0.149 & 0.0776 & 0.176 \\

$f_\ell$ IoU mean [-]
& \textbf{0.768} & 0.611 & 0.252 & 0.758 & 0.000930 \\

\bottomrule
\end{tabular}%
}
\end{table*}

\paragraph{Generalization to new materials.}
The LP-FNO framework is, in principle, material-agnostic at the architectural level; however, the trained model presented in this work is specific to Ti-6Al-4V, since only this material was considered in the training data. Generalization to other materials would therefore require either retraining or extending the input parameterization to include relevant thermophysical properties, such as density, heat capacity, thermal conductivity, absorptivity, latent heat, and solidus/liquidus temperatures. A rigorous assessment of cross-material generalization is left for future work. One possible strategy would be to define a target class of materials and generate training data spanning the corresponding range of material properties, including synthetic combinations of properties that do not necessarily correspond to a specific existing alloy. This would allow the model to learn trends across material-property space rather than memorizing isolated material cases. Open questions remain regarding the required density of training samples and the most efficient way to cover this space while minimizing the number of expensive simulations. To reduce the computational cost, a multi-fidelity strategy could be employed, in which the model is first trained on coarse-mesh, lower-fidelity simulations and subsequently fine-tuned on a smaller set of high-fidelity simulations.

\paragraph{Possible future extensions.}

Other potential immediate extensions of the present framework include coupling the LP-FNO predictions to additional process-relevant quantities beyond temperature and phase interfaces. For example, microstructure evolution, cooling rates, residual stresses, and distortion could be incorporated either as additional output fields or through a sequential surrogate pipeline linked to dedicated solidification and mechanics models. The same moving-frame and operator-learning formulation is also not limited to laser welding, and could be extended to other moving heat-source processes such as laser powder bed fusion, directed energy deposition, electron beam welding, or electron beam melting, provided that suitable high-fidelity training data are available. Another important direction is the integration of experimental data, for example metallographic melt-pool cross-sections or in-situ thermal measurements, to calibrate or fine-tune the simulation-trained surrogate and improve its predictive reliability under real processing conditions. Lastly, moving beyond the quasi-steady formulation toward transient neural-operator surrogates would allow highly unstable keyhole regimes, high-frequency interface fluctuations, pore formation, and arbitrary nonlinear welding trajectories to be modeled. However, this would constitute a substantially more challenging learning problem due to the increased dimensionality of the input-output mapping and the greater complexity of the transient thermo-fluid dynamics.

\paragraph{Long-term methodological outlook.}

Beyond these application-driven extensions, a longer-term methodological direction concerns the parameter efficiency of high-resolution 3D neural operators. An important limitation of high-resolution 3D FNO surrogates is that their dominant trainable component, the dense spectral channel-mixing weights, can become a parameter and memory bottleneck as the number of retained modes and latent channels increases, especially when extending to richer physics or higher-fidelity meshes. A natural future direction is therefore to explore parameter-efficient spectral mixing, including hybrid quantum-classical variants where a configurable subset of Fourier-layer channel mixing is produced by a small, mode-shared parametrized quantum circuit rather than many independent dense matrices. The motivation for quantum circuits here is not near-term runtime acceleration, but representational efficiency. 

\section{Conclusion}

In this work, we presented LP-FNO, a Fourier Neural Operator surrogate model for predicting quasi-steady melt-pool characteristics in laser processing across regimes ranging from conduction to stable keyhole welding. High-fidelity thermo-fluid simulations generated with FLOW-3D WELD\textsuperscript{\textregistered} were used as reference data, and the transient laser-processing problem was reformulated in a moving laser frame. By applying temporal averaging after the melt pool had formed, the problem was recast into a quasi-steady operator-learning setting suitable for predicting three-dimensional temperature fields and phase-interface geometries directly from process parameters. This quasi-steady formulation is best suited to conduction and stable keyhole regimes; highly unstable keyhole collapse, pore formation, and strongly transient welding paths would require a transient surrogate formulation.

The trained LP-FNO achieved accurate predictions of temperature, metal--gas interface, and melt-pool geometry over the investigated Ti-6Al-4V process window, while reducing inference time from minutes or hours for high-fidelity simulations to milliseconds. Compared with the tested FC-NN, U-Net, and DeepONet architectures, LP-FNO provided the most favorable overall trade-off between accuracy, inference speed, and resolution-invariant evaluation as seen in Tables \ref{tab:runtime_comparison}, \ref{tab:model_metrics_comparison}, and \ref{tab:metrics_superresolved_comparison}. The FC-NN baseline was nevertheless competitive for several temperature and interface metrics, indicating that coordinate-based models can capture a substantial part of the process-parameter dependence, although at significantly higher inference cost on large grids.

A key advantage of LP-FNO is its ability to evaluate the learned operator on grids finer than those used during training. The super-resolution results show that this capability is effective when the coarse training data already contain the relevant physical structure, as observed in the conduction regime. In the keyhole regime, however, discrepancies between coarse- and fine-mesh FLOW-3D WELD\textsuperscript{\textregistered} simulations limit the achievable fidelity of super-resolved predictions. This highlights that neural-operator super-resolution is ultimately bounded by the physical fidelity and mesh convergence of the training data.

Future work should focus on experimental calibration and validation, multi-material training, coupling to additional physics such as microstructure evolution and residual stress, extension beyond the quasi-steady formulation toward transient neural-operator surrogates for highly unstable keyhole regimes and nonlinear welding trajectories, and more parameter-efficient adaptations of the FNO architecture.

\section*{Statements and Declarations}

The authors declare that they have no competing interests. We would like to thank Empa and Terra Quantum AG for supporting this work. 
\noindent

\bigskip

\bibliography{combined_bib_deduplicated}


\begin{thebibliography}{43}
\ifx \bisbn   \undefined \def \bisbn  #1{ISBN #1}\fi
\ifx \binits  \undefined \def \binits#1{#1}\fi
\ifx \bauthor  \undefined \def \bauthor#1{#1}\fi
\ifx \batitle  \undefined \def \batitle#1{#1}\fi
\ifx \bjtitle  \undefined \def \bjtitle#1{#1}\fi
\ifx \bvolume  \undefined \def \bvolume#1{\textbf{#1}}\fi
\ifx \byear  \undefined \def \byear#1{#1}\fi
\ifx \bissue  \undefined \def \bissue#1{#1}\fi
\ifx \bfpage  \undefined \def \bfpage#1{#1}\fi
\ifx \blpage  \undefined \def \blpage #1{#1}\fi
\ifx \burl  \undefined \def \burl#1{\textsf{#1}}\fi
\ifx \doiurl  \undefined \def \doiurl#1{\url{https://doi.org/#1}}\fi
\ifx \betal  \undefined \def \betal{\textit{et al.}}\fi
\ifx \binstitute  \undefined \def \binstitute#1{#1}\fi
\ifx \binstitutionaled  \undefined \def \binstitutionaled#1{#1}\fi
\ifx \bctitle  \undefined \def \bctitle#1{#1}\fi
\ifx \beditor  \undefined \def \beditor#1{#1}\fi
\ifx \bpublisher  \undefined \def \bpublisher#1{#1}\fi
\ifx \bbtitle  \undefined \def \bbtitle#1{#1}\fi
\ifx \bedition  \undefined \def \bedition#1{#1}\fi
\ifx \bseriesno  \undefined \def \bseriesno#1{#1}\fi
\ifx \blocation  \undefined \def \blocation#1{#1}\fi
\ifx \bsertitle  \undefined \def \bsertitle#1{#1}\fi
\ifx \bsnm \undefined \def \bsnm#1{#1}\fi
\ifx \bsuffix \undefined \def \bsuffix#1{#1}\fi
\ifx \bparticle \undefined \def \bparticle#1{#1}\fi
\ifx \barticle \undefined \def \barticle#1{#1}\fi
\bibcommenthead
\ifx \bconfdate \undefined \def \bconfdate #1{#1}\fi
\ifx \botherref \undefined \def \botherref #1{#1}\fi
\ifx \url \undefined \def \url#1{\textsf{#1}}\fi
\ifx \bchapter \undefined \def \bchapter#1{#1}\fi
\ifx \bbook \undefined \def \bbook#1{#1}\fi
\ifx \bcomment \undefined \def \bcomment#1{#1}\fi
\ifx \oauthor \undefined \def \oauthor#1{#1}\fi
\ifx \citeauthoryear \undefined \def \citeauthoryear#1{#1}\fi
\ifx \endbibitem  \undefined \def \endbibitem {}\fi
\ifx \bconflocation  \undefined \def \bconflocation#1{#1}\fi
\ifx \arxivurl  \undefined \def \arxivurl#1{\textsf{#1}}\fi
\csname PreBibitemsHook\endcsname

\bibitem[\protect\citeauthoryear{Ahrens et~al.}{2005}]{ahrens_paraview_2005}
\begin{botherref}
\oauthor{\bsnm{Ahrens}, \binits{J.}},
\oauthor{\bsnm{Geveci}, \binits{B.}},
\oauthor{\bsnm{Law}, \binits{C.}}:
{ParaView}: {An} {End}-{User} {Tool} for {Large}-{Data} {Visualization}.
Elsevier
(2005).
\doiurl{10.1016/B978-012387582-2/50038-1} .
\url{https://linkinghub.elsevier.com/retrieve/pii/B9780123875822500381}
Accessed 2026-01-27
\end{botherref}
\endbibitem

\bibitem[\protect\citeauthoryear{Ahn et~al.}{2025}]{ahn_novel_2025}
\begin{barticle}
\bauthor{\bsnm{Ahn}, \binits{S.-H.}},
\bauthor{\bsnm{Song}, \binits{I.}},
\bauthor{\bsnm{Bae}, \binits{J.}},
\bauthor{\bsnm{Jeong}, \binits{G.-H.}},
\bauthor{\bsnm{Cho}, \binits{D.-W.}},
\bauthor{\bsnm{Park}, \binits{Y.W.}}:
\batitle{A novel heat source model for welding and additive manufacturing:
  {The} traveling steady pool approach}.
\bjtitle{Journal of Manufacturing Processes}
\bvolume{153},
\bfpage{471}--\blpage{486}
(\byear{2025})
\doiurl{10.1016/j.jmapro.2025.09.012}
\end{barticle}
\endbibitem

\bibitem[\protect\citeauthoryear{Aggarwal
  et~al.}{2024}]{aggarwal2024unravelling}
\begin{barticle}
\bauthor{\bsnm{Aggarwal}, \binits{A.}},
\bauthor{\bsnm{Shin}, \binits{Y.C.}},
\bauthor{\bsnm{Kumar}, \binits{A.}}:
\batitle{Unravelling keyhole instabilities and laser absorption dynamics during
  laser irradiation of {Ti6Al4V}: A high-fidelity thermo-fluidic study}.
\bjtitle{International Journal of Heat and Mass Transfer}
\bvolume{219},
\bfpage{124841}
(\byear{2024})
\doiurl{10.1016/j.ijheatmasstransfer.2023.124841}
\end{barticle}
\endbibitem

\bibitem[\protect\citeauthoryear{Bischof and
  Kraus}{2025}]{bischof_multi-objective_2025}
\begin{barticle}
\bauthor{\bsnm{Bischof}, \binits{R.}},
\bauthor{\bsnm{Kraus}, \binits{M.A.}}:
\batitle{Multi-{Objective} {Loss} {Balancing} for {Physics}-{Informed} {Deep}
  {Learning}}.
\bjtitle{Computer Methods in Applied Mechanics and Engineering}
\bvolume{439},
\bfpage{117914}
(\byear{2025})
\doiurl{10.1016/j.cma.2025.117914} .
Accessed 2026-01-25
\end{barticle}
\endbibitem

\bibitem[\protect\citeauthoryear{Bayat et~al.}{2024}]{bayat_exploring_2024}
\begin{barticle}
\bauthor{\bsnm{Bayat}, \binits{M.}},
\bauthor{\bsnm{Rothfelder}, \binits{R.}},
\bauthor{\bsnm{Schwarzkopf}, \binits{K.}},
\bauthor{\bsnm{Zinoviev}, \binits{A.}},
\bauthor{\bsnm{Zinovieva}, \binits{O.}},
\bauthor{\bsnm{Spurk}, \binits{C.}},
\bauthor{\bsnm{Hummel}, \binits{M.}},
\bauthor{\bsnm{Olowinsky}, \binits{A.}},
\bauthor{\bsnm{Beckmann}, \binits{F.}},
\bauthor{\bsnm{Moosmann}, \binits{J.}},
\bauthor{\bsnm{Schmidt}, \binits{M.}},
\bauthor{\bsnm{Hattel}, \binits{J.H.}}:
\batitle{Exploring spatial beam shaping in laser powder bed fusion:
  {High}-fidelity simulation and in-situ monitoring}.
\bjtitle{Additive Manufacturing}
\bvolume{93},
\bfpage{104420}
(\byear{2024})
\doiurl{10.1016/j.addma.2024.104420}
\end{barticle}
\endbibitem

\bibitem[\protect\citeauthoryear{Bayat
  et~al.}{2019}]{bayat_keyhole-induced_2019}
\begin{barticle}
\bauthor{\bsnm{Bayat}, \binits{M.}},
\bauthor{\bsnm{Thanki}, \binits{A.}},
\bauthor{\bsnm{Mohanty}, \binits{S.}},
\bauthor{\bsnm{Witvrouw}, \binits{A.}},
\bauthor{\bsnm{Yang}, \binits{S.}},
\bauthor{\bsnm{Thorborg}, \binits{J.}},
\bauthor{\bsnm{Tiedje}, \binits{N.S.}},
\bauthor{\bsnm{Hattel}, \binits{J.H.}}:
\batitle{Keyhole-induced porosities in {Laser}-based {Powder} {Bed} {Fusion}
  ({L}-{PBF}) of {Ti6Al4V}: {High}-fidelity modelling and experimental
  validation}.
\bjtitle{Additive Manufacturing}
\bvolume{30},
\bfpage{100835}
(\byear{2019})
\doiurl{10.1016/j.addma.2019.100835}
\end{barticle}
\endbibitem

\bibitem[\protect\citeauthoryear{Chen et~al.}{2023}]{chen_symbolic_2023}
\begin{bchapter}
\bauthor{\bsnm{Chen}, \binits{X.}},
\bauthor{\bsnm{Liang}, \binits{C.}},
\bauthor{\bsnm{Huang}, \binits{D.}},
\bauthor{\bsnm{Real}, \binits{E.}},
\bauthor{\bsnm{Wang}, \binits{K.}},
\bauthor{\bsnm{Pham}, \binits{H.}},
\bauthor{\bsnm{Dong}, \binits{X.}},
\bauthor{\bsnm{Luong}, \binits{T.}},
\bauthor{\bsnm{Hsieh}, \binits{C.-J.}},
\bauthor{\bsnm{Lu}, \binits{Y.}},
\bauthor{\bsnm{Le}, \binits{Q.V.}}:
\bctitle{Symbolic discovery of optimization algorithms}.
In: \bbtitle{Advances in Neural Information Processing Systems},
vol. \bseriesno{36}
(\byear{2023}).
\doiurl{10.52202/075280-2140}
\end{bchapter}
\endbibitem

\bibitem[\protect\citeauthoryear{Chen et~al.}{2023}]{chen_numerical_2023}
\begin{barticle}
\bauthor{\bsnm{Chen}, \binits{D.}},
\bauthor{\bsnm{Li}, \binits{G.}},
\bauthor{\bsnm{Wang}, \binits{P.}},
\bauthor{\bsnm{Zeng}, \binits{Z.}},
\bauthor{\bsnm{Tang}, \binits{Y.}}:
\batitle{Numerical simulation of melt pool size and flow evolution for laser
  powder bed fusion of powder grade {Ti6Al4V}}.
\bjtitle{Finite Elements in Analysis and Design}
\bvolume{223},
\bfpage{103971}
(\byear{2023})
\doiurl{10.1016/j.finel.2023.103971}
\end{barticle}
\endbibitem

\bibitem[\protect\citeauthoryear{De and DebRoy}{2004}]{de_smart_2004}
\begin{barticle}
\bauthor{\bsnm{De}, \binits{A.}},
\bauthor{\bsnm{DebRoy}, \binits{T.}}:
\batitle{A smart model to estimate effective thermal conductivity and viscosity
  in the weld pool}.
\bjtitle{Journal of Applied Physics}
\bvolume{95}(\bissue{9}),
\bfpage{5230}--\blpage{5240}
(\byear{2004})
\doiurl{10.1063/1.1695593} .
Accessed 2026-01-26
\end{barticle}
\endbibitem

\bibitem[\protect\citeauthoryear{Fulco et~al.}{2026}]{fulco_numerical_2026}
\begin{barticle}
\bauthor{\bsnm{Fulco}, \binits{E.}},
\bauthor{\bsnm{Coviello}, \binits{D.}},
\bauthor{\bsnm{Sargente}, \binits{D.}}:
\batitle{Numerical analysis of the effect of stationary laser beam properties
  on {Al}-{Si} coating mixing in {22MnB5} steel}.
\bjtitle{International Journal of Heat and Mass Transfer}
\bvolume{255}(\bissue{2}),
\bfpage{127899}
(\byear{2026})
\doiurl{10.1016/j.ijheatmasstransfer.2025.127899}
\end{barticle}
\endbibitem

\bibitem[\protect\citeauthoryear{{Flow Science,
  Inc.}}{2025}]{noauthor_flow-3d_2025}
\begin{botherref}
\oauthor{\bsnm{{Flow Science, Inc.}}}:
{FLOW}-{3D} {WELD}®.
Flow Science, Inc.,
Santa Fe, NM, USA
(2025).
\url{https://flow3d.com}
\end{botherref}
\endbibitem

\bibitem[\protect\citeauthoryear{Flint et~al.}{2023}]{flint_laserbeamfoam_2023}
\begin{barticle}
\bauthor{\bsnm{Flint}, \binits{T.F.}},
\bauthor{\bsnm{Robson}, \binits{J.D.}},
\bauthor{\bsnm{Parivendhan}, \binits{G.}},
\bauthor{\bsnm{Cardiff}, \binits{P.}}:
\batitle{{laserbeamFoam}: {Laser} ray-tracing and thermally induced state
  transition simulation toolkit}.
\bjtitle{SoftwareX}
\bvolume{21},
\bfpage{101299}
(\byear{2023})
\doiurl{10.1016/j.softx.2022.101299} .
Accessed 2025-09-16
\end{barticle}
\endbibitem

\bibitem[\protect\citeauthoryear{Hendrycks and
  Gimpel}{2016}]{hendrycks_gaussian_2016}
\begin{botherref}
\oauthor{\bsnm{Hendrycks}, \binits{D.}},
\oauthor{\bsnm{Gimpel}, \binits{K.}}:
Gaussian {Error} {Linear} {Units} ({GELUs}).
arXiv.
Version Number: 5
(2016).
\doiurl{10.48550/ARXIV.1606.08415} .
\url{https://arxiv.org/abs/1606.08415}
Accessed 2026-01-25
\end{botherref}
\endbibitem

\bibitem[\protect\citeauthoryear{Hann et~al.}{2011}]{hann_simple_2011}
\begin{barticle}
\bauthor{\bsnm{Hann}, \binits{D.B.}},
\bauthor{\bsnm{Iammi}, \binits{J.}},
\bauthor{\bsnm{Folkes}, \binits{J.}}:
\batitle{A simple methodology for predicting laser-weld properties from
  material and laser parameters}.
\bjtitle{Journal of Physics D: Applied Physics}
\bvolume{44}(\bissue{44}),
\bfpage{445401}
(\byear{2011})
\doiurl{10.1088/0022-3727/44/44/445401} .
Accessed 2026-01-07
\end{barticle}
\endbibitem

\bibitem[\protect\citeauthoryear{Hemmasian
  et~al.}{2022}]{hemmasian_surrogate_2022}
\begin{botherref}
\oauthor{\bsnm{Hemmasian}, \binits{A.}},
\oauthor{\bsnm{Ogoke}, \binits{O.F.}},
\oauthor{\bsnm{Akbari}, \binits{P.}},
\oauthor{\bsnm{Malen}, \binits{J.}},
\oauthor{\bsnm{Beuth}, \binits{J.}},
\oauthor{\bsnm{Barati~Farimani}, \binits{A.}}:
Surrogate {Modeling} of {Melt} {Pool} {Thermal} {Field} {Using} {Deep}
  {Learning}.
Social Science Research Network,
Rochester, NY
(2022).
\doiurl{10.2139/ssrn.4190835} .
\url{https://papers.ssrn.com/abstract=4190835}
Accessed 2026-01-25
\end{botherref}
\endbibitem

\bibitem[\protect\citeauthoryear{Kurz et~al.}{2023}]{kurz_fundamentals_2023}
\begin{bbook}
\bauthor{\bsnm{Kurz}, \binits{W.}},
\bauthor{\bsnm{Fisher}, \binits{D.J.}},
\bauthor{\bsnm{Rappaz}, \binits{M.}}:
\bbtitle{Fundamentals of Solidification},
\bedition{5th fully revised edition} edn.
\bsertitle{Foundations of Materials Science and Engineering},
vol. \bseriesno{103}.
\bpublisher{Trans Tech Publications Ltd},
\blocation{B{\"a}ch, Switzerland}
(\byear{2023}).
\doiurl{10.4028/b-v7b35q} .
\burl{https://infoscience.epfl.ch/handle/20.500.14299/245295}
\end{bbook}
\endbibitem

\bibitem[\protect\citeauthoryear{Karniadakis
  et~al.}{2021}]{karniadakis_physics-informed_2021}
\begin{barticle}
\bauthor{\bsnm{Karniadakis}, \binits{G.E.}},
\bauthor{\bsnm{Kevrekidis}, \binits{I.G.}},
\bauthor{\bsnm{Lu}, \binits{L.}},
\bauthor{\bsnm{Perdikaris}, \binits{P.}},
\bauthor{\bsnm{Wang}, \binits{S.}},
\bauthor{\bsnm{Yang}, \binits{L.}}:
\batitle{Physics-informed machine learning}.
\bjtitle{Nature Reviews Physics}
\bvolume{3}(\bissue{6}),
\bfpage{422}--\blpage{440}
(\byear{2021})
\doiurl{10.1038/s42254-021-00314-5} .
Accessed 2026-01-25
\end{barticle}
\endbibitem

\bibitem[\protect\citeauthoryear{Ko et~al.}{2024}]{ko_review_2024}
\begin{barticle}
\bauthor{\bsnm{Ko}, \binits{T.}},
\bauthor{\bsnm{Kim}, \binits{H.}},
\bauthor{\bsnm{Shin}, \binits{Y.}},
\bauthor{\bsnm{Kim}, \binits{D.}},
\bauthor{\bsnm{Lee}, \binits{Y.H.}},
\bauthor{\bsnm{Hong}, \binits{J.}},
\bauthor{\bsnm{Lee}, \binits{S.H.}}:
\batitle{Review of {Recent} {Additive} {Manufacturing} and {Welding} {Research}
  with {Application} of {Physics}-{Informed} {Neural} {Networks}}.
\bjtitle{Journal of Welding and Joining}
\bvolume{42}(\bissue{4}),
\bfpage{357}--\blpage{365}
(\byear{2024})
\doiurl{10.5781/JWJ.2024.42.4.3} .
Accessed 2025-10-22
\end{barticle}
\endbibitem

\bibitem[\protect\citeauthoryear{Kovachki et~al.}{2023}]{kovachki_neural_2023}
\begin{barticle}
\bauthor{\bsnm{Kovachki}, \binits{N.}},
\bauthor{\bsnm{Li}, \binits{Z.}},
\bauthor{\bsnm{Liu}, \binits{B.}},
\bauthor{\bsnm{Azizzadenesheli}, \binits{K.}},
\bauthor{\bsnm{Bhattacharya}, \binits{K.}},
\bauthor{\bsnm{Stuart}, \binits{A.}},
\bauthor{\bsnm{Anandkumar}, \binits{A.}}:
\batitle{Neural {Operator}: {Learning} {Maps} {Between} {Function} {Spaces}
  {With} {Applications} to {PDEs}}.
\bjtitle{Journal of Machine Learning Research}
\bvolume{24}(\bissue{89}),
\bfpage{1}--\blpage{97}
(\byear{2023}).
\bcomment{\url{{http://jmlr.org/papers/v24/21-1524.html}}}
\end{barticle}
\endbibitem

\bibitem[\protect\citeauthoryear{Katinas and
  Shin}{2020}]{katinas2020prediction}
\begin{barticle}
\bauthor{\bsnm{Katinas}, \binits{C.}},
\bauthor{\bsnm{Shin}, \binits{Y.C.}}:
\batitle{Prediction of initial transient behavior with stationary heating
  during laser powder bed fusion processes}.
\bjtitle{International Journal of Heat and Mass Transfer}
\bvolume{153},
\bfpage{119663}
(\byear{2020})
\doiurl{10.1016/j.ijheatmasstransfer.2020.119663}
\end{barticle}
\endbibitem

\bibitem[\protect\citeauthoryear{Kamara et~al.}{2011}]{kamara_modelling_2011}
\begin{barticle}
\bauthor{\bsnm{Kamara}, \binits{A.M.}},
\bauthor{\bsnm{Wang}, \binits{W.}},
\bauthor{\bsnm{Marimuthu}, \binits{S.}},
\bauthor{\bsnm{Li}, \binits{L.}}:
\batitle{Modelling of the {Melt} {Pool} {Geometry} in the {Laser} {Deposition}
  of {Nickel} {Alloys} {Using} the {Anisotropic} {Enhanced} {Thermal}
  {Conductivity} {Approach}}.
\bjtitle{Proceedings of the Institution of Mechanical Engineers, Part B:
  Journal of Engineering Manufacture}
\bvolume{225}(\bissue{1}),
\bfpage{87}--\blpage{99}
(\byear{2011})
\doiurl{10.1177/09544054JEM2129} .
Accessed 2026-01-26
\end{barticle}
\endbibitem

\bibitem[\protect\citeauthoryear{Li et~al.}{2023}]{li_fourier_2023}
\begin{barticle}
\bauthor{\bsnm{Li}, \binits{Z.}},
\bauthor{\bsnm{Huang}, \binits{D.Z.}},
\bauthor{\bsnm{Liu}, \binits{B.}}:
\batitle{Fourier {Neural} {Operator} with {Learned} {Deformations} for {PDEs}
  on {General} {Geometries}}.
\bjtitle{Journal of Machine Learning Research}
\bvolume{24}(\bissue{388}),
\bfpage{1}--\blpage{26}
(\byear{2023}).
\bcomment{\url{http://www.jmlr.org/papers/v24/23-0064.html}}
\end{barticle}
\endbibitem

\bibitem[\protect\citeauthoryear{Lu et~al.}{2021}]{lu_deeponet_2021}
\begin{barticle}
\bauthor{\bsnm{Lu}, \binits{L.}},
\bauthor{\bsnm{Jin}, \binits{P.}},
\bauthor{\bsnm{Karniadakis}, \binits{G.E.}}:
\batitle{{DeepONet}: {Learning} nonlinear operators for identifying
  differential equations based on the universal approximation theorem of
  operators}.
\bjtitle{Nature Machine Intelligence}
\bvolume{3}(\bissue{3}),
\bfpage{218}--\blpage{229}
(\byear{2021})
\doiurl{10.1038/s42256-021-00302-5} .
Accessed 2025-09-04
\end{barticle}
\endbibitem

\bibitem[\protect\citeauthoryear{Li et~al.}{2021}]{li_fourier_2021}
\begin{bchapter}
\bauthor{\bsnm{Li}, \binits{Z.}},
\bauthor{\bsnm{Kovachki}, \binits{N.}},
\bauthor{\bsnm{Azizzadenesheli}, \binits{K.}},
\bauthor{\bsnm{Liu}, \binits{B.}},
\bauthor{\bsnm{Bhattacharya}, \binits{K.}},
\bauthor{\bsnm{Stuart}, \binits{A.}},
\bauthor{\bsnm{Anandkumar}, \binits{A.}}:
\bctitle{Fourier neural operator for parametric partial differential
  equations}.
In: \bbtitle{International Conference on Learning Representations}
(\byear{2021}).
\burl{https://openreview.net/forum?id=c8P9NQVtmnO}
\end{bchapter}
\endbibitem

\bibitem[\protect\citeauthoryear{Liu et~al.}{2024}]{liu_deep_2024}
\begin{barticle}
\bauthor{\bsnm{Liu}, \binits{N.}},
\bauthor{\bsnm{Li}, \binits{X.}},
\bauthor{\bsnm{Rajanna}, \binits{M.R.}},
\bauthor{\bsnm{Reutzel}, \binits{E.W.}},
\bauthor{\bsnm{Sawyer}, \binits{B.}},
\bauthor{\bsnm{Rao}, \binits{P.}},
\bauthor{\bsnm{Lua}, \binits{J.}},
\bauthor{\bsnm{Phan}, \binits{N.}},
\bauthor{\bsnm{Yu}, \binits{Y.}}:
\batitle{Deep neural operator enabled digital twin modeling for additive
  manufacturing}.
\bjtitle{Advances in Computational Science and Engineering}
\bvolume{2}(\bissue{3}),
\bfpage{174}--\blpage{201}
(\byear{2024})
\doiurl{10.3934/acse.2024010} .
Accessed 2025-11-18
\end{barticle}
\endbibitem

\bibitem[\protect\citeauthoryear{Li et~al.}{2024}]{li_physics-informed_2024}
\begin{barticle}
\bauthor{\bsnm{Li}, \binits{Z.}},
\bauthor{\bsnm{Zheng}, \binits{H.}},
\bauthor{\bsnm{Kovachki}, \binits{N.}},
\bauthor{\bsnm{Jin}, \binits{D.}},
\bauthor{\bsnm{Chen}, \binits{H.}},
\bauthor{\bsnm{Liu}, \binits{B.}},
\bauthor{\bsnm{Azizzadenesheli}, \binits{K.}},
\bauthor{\bsnm{Anandkumar}, \binits{A.}}:
\batitle{Physics-{Informed} {Neural} {Operator} for {Learning} {Partial}
  {Differential} {Equations}}.
\bjtitle{ACM / IMS Journal of Data Science}
\bvolume{1}(\bissue{3}),
\bfpage{1}--\blpage{27}
(\byear{2024})
\doiurl{10.1145/3648506} .
Accessed 2025-09-04
\end{barticle}
\endbibitem

\bibitem[\protect\citeauthoryear{Moreira
  et~al.}{2025}]{moreira_high-fidelity_2025}
\begin{barticle}
\bauthor{\bsnm{Moreira}, \binits{C.A.}},
\bauthor{\bsnm{Chiumenti}, \binits{M.}},
\bauthor{\bsnm{Caicedo}, \binits{M.A.}},
\bauthor{\bsnm{Baiges}, \binits{J.}},
\bauthor{\bsnm{Cervera}, \binits{M.}}:
\batitle{High-fidelity part-scale simulations in metal additive manufacturing
  using a computationally efficient and accurate approach}.
\bjtitle{Additive Manufacturing}
\bvolume{104},
\bfpage{104748}
(\byear{2025})
\doiurl{10.1016/j.addma.2025.104748} .
Accessed 2026-01-25
\end{barticle}
\endbibitem

\bibitem[\protect\citeauthoryear{Michaleris}{2014}]{michaleris_modeling_2014}
\begin{barticle}
\bauthor{\bsnm{Michaleris}, \binits{P.}}:
\batitle{Modeling metal deposition in heat transfer analyses of additive
  manufacturing processes}.
\bjtitle{Finite Elements in Analysis and Design}
\bvolume{86},
\bfpage{51}--\blpage{60}
(\byear{2014})
\doiurl{10.1016/j.finel.2014.04.003} .
Accessed 2026-01-25
\end{barticle}
\endbibitem

\bibitem[\protect\citeauthoryear{Manav et~al.}{2025}]{manav_meltpoolinr_2025}
\begin{barticle}
\bauthor{\bsnm{Manav}, \binits{M.}},
\bauthor{\bsnm{Perraudin}, \binits{N.}},
\bauthor{\bsnm{Lin}, \binits{Y.}},
\bauthor{\bsnm{Afrasiabi}, \binits{M.}},
\bauthor{\bsnm{Perez-Cruz}, \binits{F.}},
\bauthor{\bsnm{Bambach}, \binits{M.}},
\bauthor{\bsnm{De~Lorenzis}, \binits{L.}}:
\batitle{{MeltpoolINR}: predicting temperature field, melt pool geometry and
  their rate of change in laser powder bed fusion}.
\bjtitle{Journal of Intelligent Manufacturing}
(\byear{2025})
\doiurl{10.1007/s10845-025-02692-4} .
Accessed 2026-01-08
\end{barticle}
\endbibitem

\bibitem[\protect\citeauthoryear{Mukherjee et~al.}{2018}]{mukherjee_heat_2018}
\begin{barticle}
\bauthor{\bsnm{Mukherjee}, \binits{T.}},
\bauthor{\bsnm{Wei}, \binits{H.L.}},
\bauthor{\bsnm{De}, \binits{A.}},
\bauthor{\bsnm{DebRoy}, \binits{T.}}:
\batitle{Heat and fluid flow in additive manufacturing—{Part} {I}: {Modeling}
  of powder bed fusion}.
\bjtitle{Computational Materials Science}
\bvolume{150},
\bfpage{304}--\blpage{313}
(\byear{2018})
\doiurl{10.1016/j.commatsci.2018.04.022} .
Accessed 2026-01-25
\end{barticle}
\endbibitem

\bibitem[\protect\citeauthoryear{Mukherjee
  et~al.}{2017}]{mukherjee_improved_2017}
\begin{barticle}
\bauthor{\bsnm{Mukherjee}, \binits{T.}},
\bauthor{\bsnm{Zhang}, \binits{W.}},
\bauthor{\bsnm{DebRoy}, \binits{T.}}:
\batitle{An improved prediction of residual stresses and distortion in additive
  manufacturing}.
\bjtitle{Computational Materials Science}
\bvolume{126},
\bfpage{360}--\blpage{372}
(\byear{2017})
\doiurl{10.1016/j.commatsci.2016.10.003} .
Accessed 2026-01-25
\end{barticle}
\endbibitem

\bibitem[\protect\citeauthoryear{{NVIDIA PhysicsNeMo
  Contributors}}{2023}]{physicsnemo_contributors_nvidia_2023}
\begin{botherref}
\oauthor{\bsnm{{NVIDIA PhysicsNeMo Contributors}}}:
{NVIDIA} {PhysicsNeMo}: {An} open-source framework for physics-based deep
  learning in science and engineering
(2023).
\url{https://github.com/NVIDIA/physicsnemo}
\end{botherref}
\endbibitem

\bibitem[\protect\citeauthoryear{Ronneberger
  et~al.}{2015}]{ronneberger_u-net_2015}
\begin{bchapter}
\bauthor{\bsnm{Ronneberger}, \binits{O.}},
\bauthor{\bsnm{Fischer}, \binits{P.}},
\bauthor{\bsnm{Brox}, \binits{T.}}:
\bctitle{U-{Net}: {Convolutional} {Networks} for {Biomedical} {Image}
  {Segmentation}}.
In: \beditor{\bsnm{Navab}, \binits{N.}},
\beditor{\bsnm{Hornegger}, \binits{J.}},
\beditor{\bsnm{Wells}, \binits{W.M.}},
\beditor{\bsnm{Frangi}, \binits{A.F.}} (eds.)
\bbtitle{Medical {Image} {Computing} and {Computer}-{Assisted} {Intervention}
  – {MICCAI} 2015}.
\bsertitle{Lecture Notes in Computer Science},
vol. \bseriesno{9351},
pp. \bfpage{234}--\blpage{241}.
\bpublisher{Springer},
\blocation{Cham}
(\byear{2015}).
\doiurl{10.1007/978-3-319-24574-4_28}
\end{bchapter}
\endbibitem

\bibitem[\protect\citeauthoryear{Raonic
  et~al.}{2023}]{raonic_convolutional_2023}
\begin{bchapter}
\bauthor{\bsnm{Raonic}, \binits{B.}},
\bauthor{\bsnm{Molinaro}, \binits{R.}},
\bauthor{\bsnm{De~Ryck}, \binits{T.}},
\bauthor{\bsnm{Rohner}, \binits{T.}},
\bauthor{\bsnm{Bartolucci}, \binits{F.}},
\bauthor{\bsnm{Alaifari}, \binits{R.}},
\bauthor{\bsnm{Mishra}, \binits{S.}},
\bauthor{\bsnm{Bezenac}, \binits{E.}}:
\bctitle{Convolutional {Neural} {Operators} for {Robust} and {Accurate}
  {Learning} of {PDEs}}.
In: \bbtitle{Advances in {Neural} {Information} {Processing} {Systems}
  ({NeurIPS}) 36}
(\byear{2023}).
\doiurl{10.48550/arXiv.2302.01178}
\end{bchapter}
\endbibitem

\bibitem[\protect\citeauthoryear{Raissi
  et~al.}{2019}]{raissi_physics-informed_2019}
\begin{barticle}
\bauthor{\bsnm{Raissi}, \binits{M.}},
\bauthor{\bsnm{Perdikaris}, \binits{P.}},
\bauthor{\bsnm{Karniadakis}, \binits{G.E.}}:
\batitle{Physics-informed neural networks: {A} deep learning framework for
  solving forward and inverse problems involving nonlinear partial differential
  equations}.
\bjtitle{Journal of Computational Physics}
\bvolume{378},
\bfpage{686}--\blpage{707}
(\byear{2019})
\doiurl{10.1016/j.jcp.2018.10.045} .
Accessed 2025-09-04
\end{barticle}
\endbibitem

\bibitem[\protect\citeauthoryear{Ransenigo
  et~al.}{2022}]{ransenigo_evolution_2022}
\begin{barticle}
\bauthor{\bsnm{Ransenigo}, \binits{C.}},
\bauthor{\bsnm{Tocci}, \binits{M.}},
\bauthor{\bsnm{Palo}, \binits{F.}},
\bauthor{\bsnm{Ginestra}, \binits{P.}},
\bauthor{\bsnm{Ceretti}, \binits{E.}},
\bauthor{\bsnm{Gelfi}, \binits{M.}},
\bauthor{\bsnm{Pola}, \binits{A.}}:
\batitle{Evolution of {Melt} {Pool} and {Porosity} {During} {Laser} {Powder}
  {Bed} {Fusion} of {Ti6Al4V} {Alloy}: {Numerical} {Modelling} and
  {Experimental} {Validation}}.
\bjtitle{Lasers in Manufacturing and Materials Processing}
\bvolume{9}(\bissue{4}),
\bfpage{481}--\blpage{502}
(\byear{2022})
\doiurl{10.1007/s40516-022-00185-3}
\end{barticle}
\endbibitem

\bibitem[\protect\citeauthoryear{Svenungsson
  et~al.}{2015}]{svenungsson2015laser}
\begin{barticle}
\bauthor{\bsnm{Svenungsson}, \binits{J.}},
\bauthor{\bsnm{Choquet}, \binits{I.}},
\bauthor{\bsnm{Kaplan}, \binits{A.F.}}:
\batitle{Laser welding process--a review of keyhole welding modelling}.
\bjtitle{Physics procedia}
\bvolume{78},
\bfpage{182}--\blpage{191}
(\byear{2015})
\doiurl{10.1016/j.phpro.2015.11.042}
\end{barticle}
\endbibitem

\bibitem[\protect\citeauthoryear{Saldi et~al.}{2013}]{saldi_effect_2013}
\begin{barticle}
\bauthor{\bsnm{Saldi}, \binits{Z.S.}},
\bauthor{\bsnm{Kidess}, \binits{A.}},
\bauthor{\bsnm{Kenjereš}, \binits{S.}},
\bauthor{\bsnm{Zhao}, \binits{C.}},
\bauthor{\bsnm{Richardson}, \binits{I.M.}},
\bauthor{\bsnm{Kleijn}, \binits{C.R.}}:
\batitle{Effect of enhanced heat and mass transport and flow reversal during
  cool down on weld pool shapes in laser spot welding of steel}.
\bjtitle{International Journal of Heat and Mass Transfer}
\bvolume{66},
\bfpage{879}--\blpage{888}
(\byear{2013})
\doiurl{10.1016/j.ijheatmasstransfer.2013.07.085} .
Accessed 2026-01-26
\end{barticle}
\endbibitem

\bibitem[\protect\citeauthoryear{Safari and
  Wessels}{2025}]{safari_physics-informed_2025}
\begin{botherref}
\oauthor{\bsnm{Safari}, \binits{H.}},
\oauthor{\bsnm{Wessels}, \binits{H.}}:
Physics-{Informed} {Surrogates} for {Temperature} {Prediction} of
  {Multi}-{Tracks} in {Laser} {Powder} {Bed} {Fusion}.
arXiv.
arXiv:2502.01820 [cs]
(2025).
\doiurl{10.48550/arXiv.2502.01820} .
\url{http://arxiv.org/abs/2502.01820}
Accessed 2025-11-18
\end{botherref}
\endbibitem

\bibitem[\protect\citeauthoryear{Torres and
  Niepert}{2025}]{torres_adaptive_2025}
\begin{botherref}
\oauthor{\bsnm{Torres}, \binits{E.}},
\oauthor{\bsnm{Niepert}, \binits{M.}}:
Adaptive physics-informed neural networks: A survey.
Transactions on Machine Learning Research
(2025).
Survey Certification
\end{botherref}
\endbibitem

\bibitem[\protect\citeauthoryear{Wu et~al.}{2018}]{wu_parametric_2018}
\begin{barticle}
\bauthor{\bsnm{Wu}, \binits{Y.-C.}},
\bauthor{\bsnm{Hwang}, \binits{W.-S.}},
\bauthor{\bsnm{San}, \binits{C.-H.}},
\bauthor{\bsnm{Chang}, \binits{C.-H.}},
\bauthor{\bsnm{Lin}, \binits{H.-J.}}:
\batitle{Parametric study of surface morphology for selective laser melting on
  {Ti6Al4V} powder bed with numerical and experimental methods}.
\bjtitle{International Journal of Material Forming}
\bvolume{11}(\bissue{6}),
\bfpage{807}--\blpage{813}
(\byear{2018})
\doiurl{10.1007/s12289-017-1391-2}
\end{barticle}
\endbibitem

\bibitem[\protect\citeauthoryear{Yaseen et~al.}{2023}]{yaseen_fast_2023}
\begin{barticle}
\bauthor{\bsnm{Yaseen}, \binits{M.}},
\bauthor{\bsnm{Yushu}, \binits{D.}},
\bauthor{\bsnm{German}, \binits{P.}},
\bauthor{\bsnm{Wu}, \binits{X.}}:
\batitle{Fast and accurate reduced-order modeling of a {MOOSE}-based additive
  manufacturing model with operator learning}.
\bjtitle{The International Journal of Advanced Manufacturing Technology}
\bvolume{129}(\bissue{7-8}),
\bfpage{3123}--\blpage{3139}
(\byear{2023})
\doiurl{10.1007/s00170-023-12471-1} .
Accessed 2026-01-05
\end{barticle}
\endbibitem

\bibitem[\protect\citeauthoryear{Zenz et~al.}{2024}]{zenz2024compressible}
\begin{barticle}
\bauthor{\bsnm{Zenz}, \binits{C.}},
\bauthor{\bsnm{Buttazzoni}, \binits{M.}},
\bauthor{\bsnm{Florian}, \binits{T.}},
\bauthor{\bsnm{Armijos}, \binits{K.E.C.}},
\bauthor{\bsnm{V{\'a}zquez}, \binits{R.G.}},
\bauthor{\bsnm{Liedl}, \binits{G.}},
\bauthor{\bsnm{Otto}, \binits{A.}}:
\batitle{A compressible multiphase mass-of-fluid model for the simulation of
  laser-based manufacturing processes}.
\bjtitle{Computers \& Fluids}
\bvolume{268},
\bfpage{106109}
(\byear{2024})
\doiurl{10.1016/j.compfluid.2023.106109}
\end{barticle}
\endbibitem

\end{thebibliography}

\newcounter{figsave}
\setcounter{figsave}{\value{figure}}
\newcounter{tabsave}
\setcounter{tabsave}{\value{table}}
\newpage
\begin{appendices}
\section{Extra figures on LP-FNO Convergence and Hyperparameters}\label{secA1}
\label{Appendix}
\setcounter{figure}{\value{figsave}}
\renewcommand{\thefigure}{\arabic{figure}}
\setcounter{table}{\value{tabsave}}
\renewcommand{\thetable}{\arabic{table}}

\begin{figure}[h]
    \centering
    \begin{subfigure}[b]{0.48\textwidth}
        \centering
        \includegraphics[width=\textwidth]{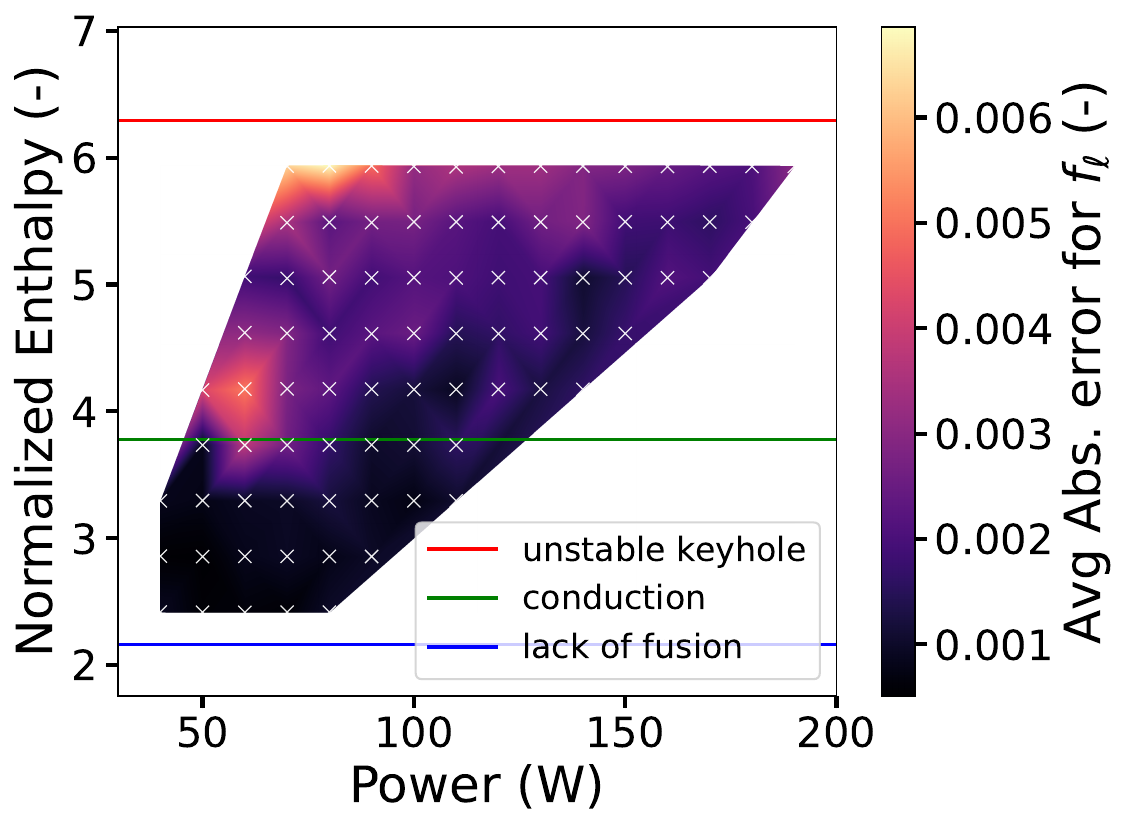}
        \caption{Avg.\ absolute error for $f_\ell$ (P--$H$)}
        \label{fig:fl_abs_err_ph}
    \end{subfigure}
    \hfill
    \begin{subfigure}[b]{0.48\textwidth}
        \centering
        \includegraphics[width=\textwidth]{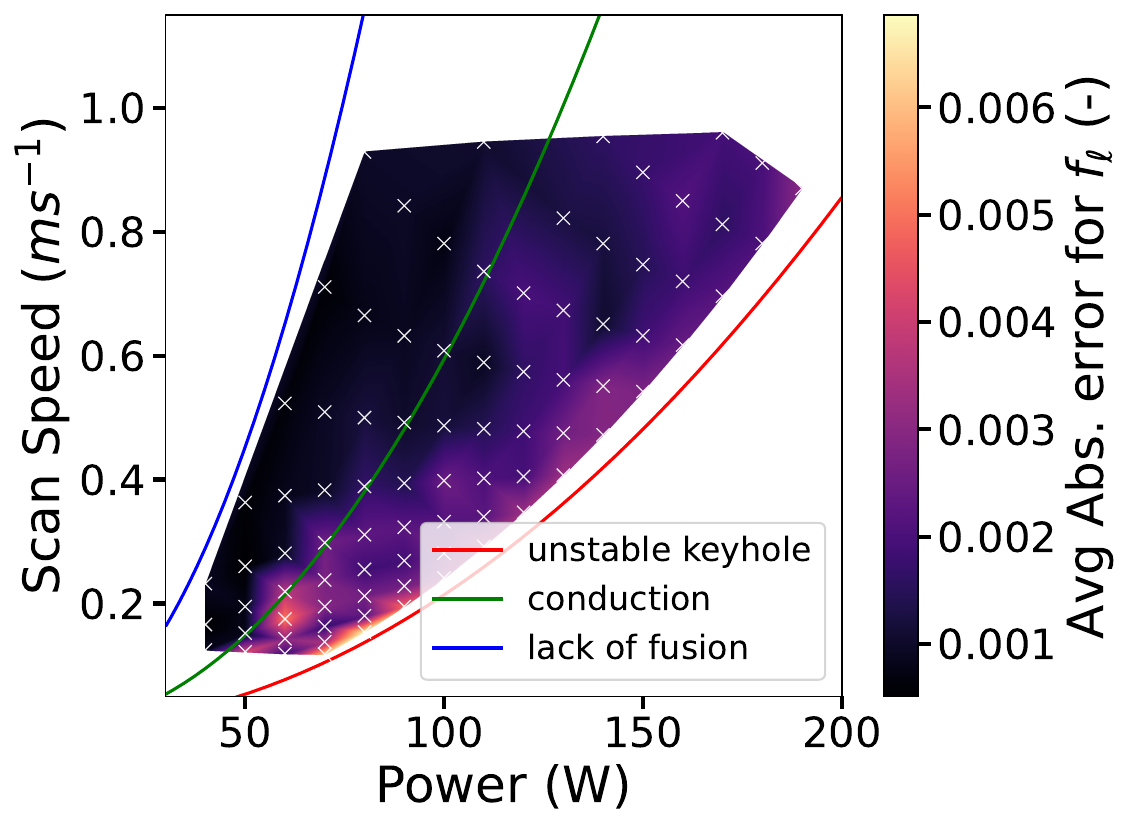}
        \caption{Avg.\ absolute error for $f_\ell$ (P--$V_{\mathrm{scan}}$)}
        \label{fig:fl_abs_err_pv}
    \end{subfigure}

    \vspace{2mm}

    \begin{subfigure}[b]{0.48\textwidth}
        \centering
        \includegraphics[width=\textwidth]{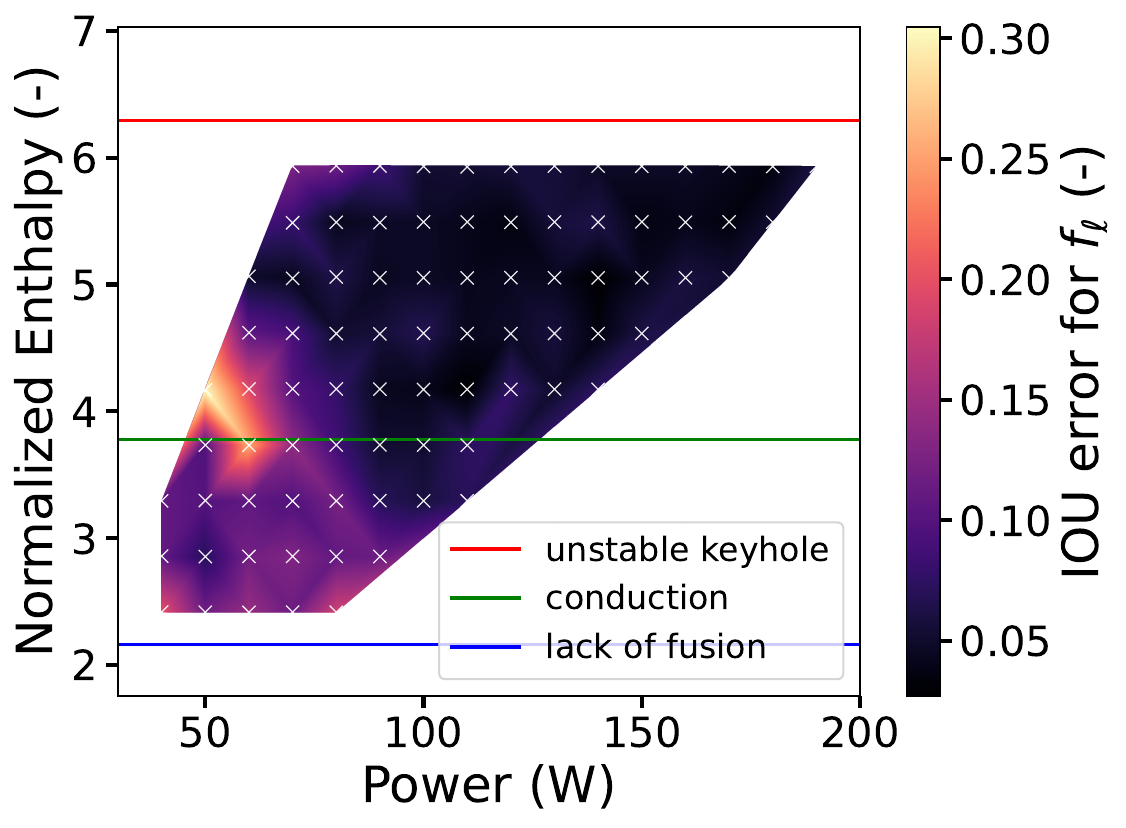}
        \caption{IoU error ($1-$IoU) for $f_\ell$ (P--$H$)}
        \label{fig:fl_iou_ph}
    \end{subfigure}
    \hfill
    \begin{subfigure}[b]{0.48\textwidth}
        \centering
        \includegraphics[width=\textwidth]{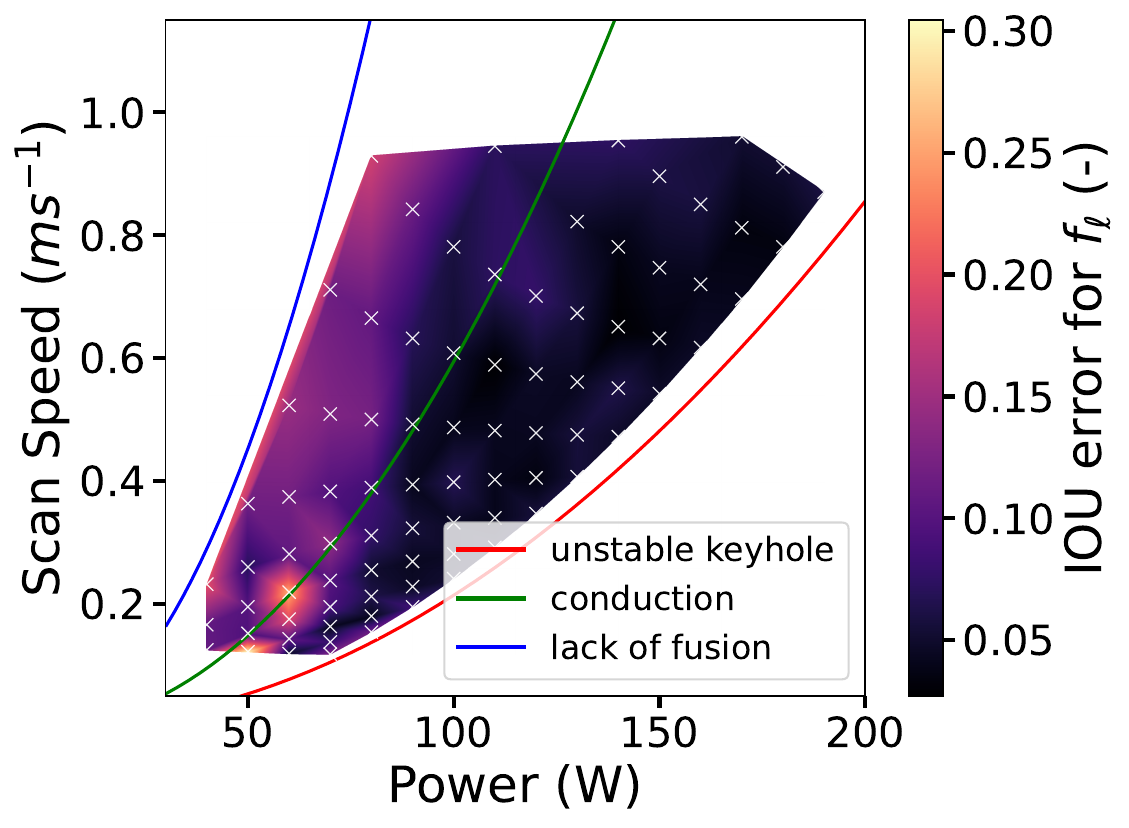}
        \caption{IoU error ($1-$IoU) for $f_\ell$ (P--$V_{\mathrm{scan}}$)}
        \label{fig:fl_iou_pv}
    \end{subfigure}

    \caption{Melt-pool (liquid fraction $f_\ell$) segmentation performance across the process window. Top row: average absolute error. Bottom row: IoU error (1-IoU) computed after thresholding at $f_\ell \ge 0.5$. Results are shown in two parameterizations: $(P,V_{\mathrm{scan}})$ and $(P,H)$. Good segmentation overall, except for one outlier at $P=\SI{50}{\watt}$, $V_{scan}=\SI{0.122}{\meter\per\second}$, which lies at edge of data in start of conduction-keyhole transition}
    \label{fig:fl_segmentation_2x2}
\end{figure}

\begin{figure}[h]
    \centering
    \begin{subfigure}[b]{0.48\textwidth}
        \centering
        \includegraphics[width=\textwidth]{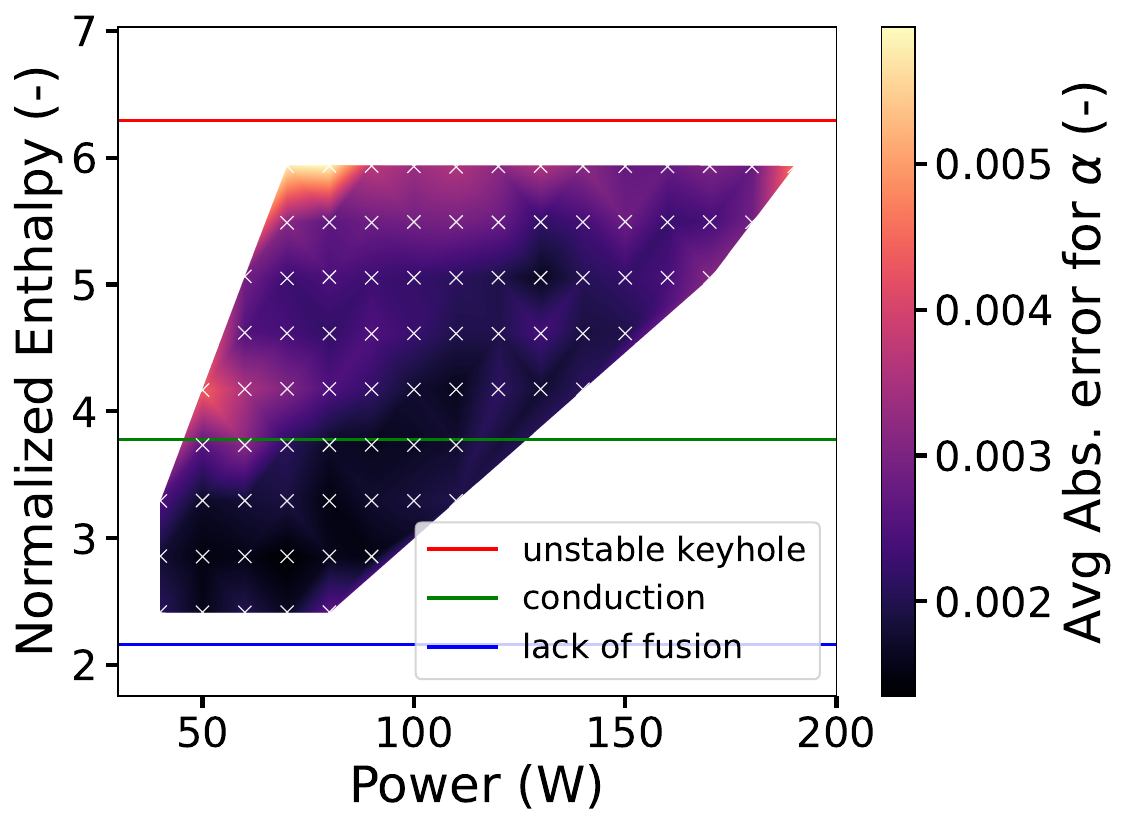}
        \caption{Avg.\ absolute error for $\alpha$ (P--$H$)}
        \label{fig:alpha_abs_err_ph}
    \end{subfigure}
    \hfill
    \begin{subfigure}[b]{0.48\textwidth}
        \centering
        \includegraphics[width=\textwidth]{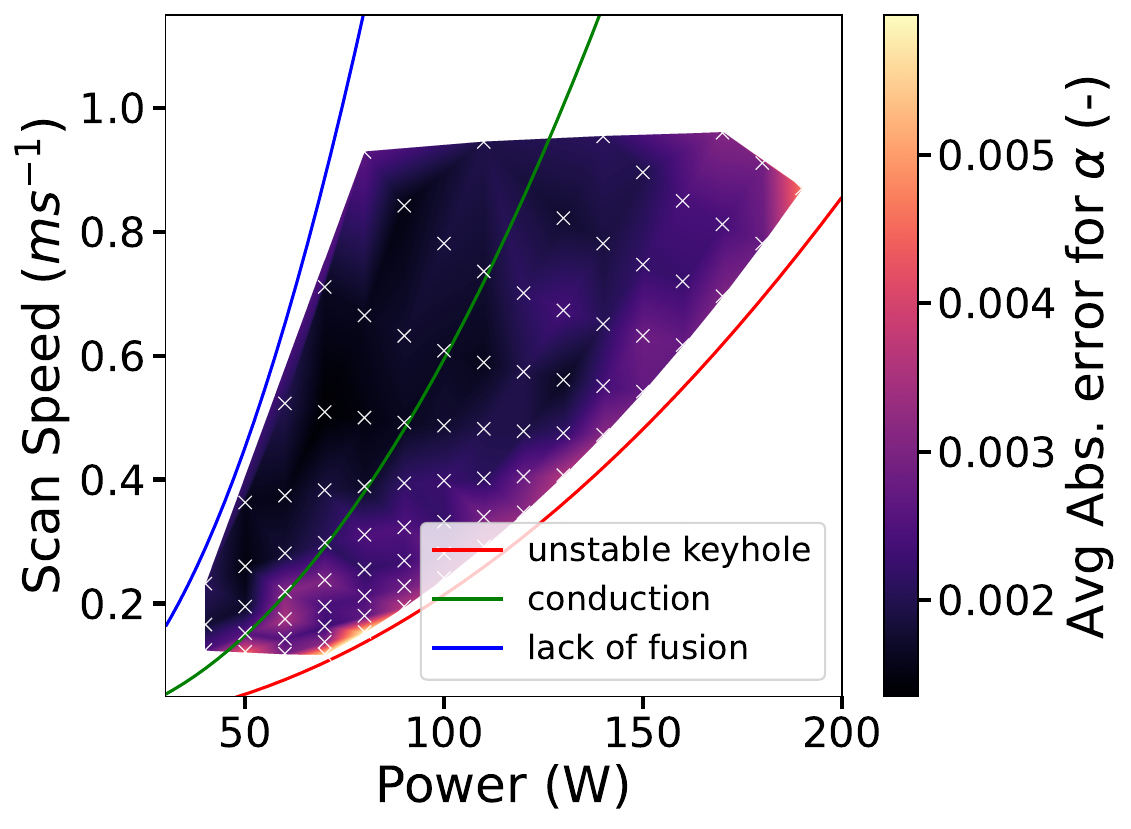}
        \caption{Avg.\ absolute error for $\alpha$ (P--$V_{\mathrm{scan}}$)}
        \label{fig:alpha_abs_err_pv}
    \end{subfigure}

    \vspace{2mm}

    \begin{subfigure}[b]{0.48\textwidth}
        \centering
        \includegraphics[width=\textwidth]{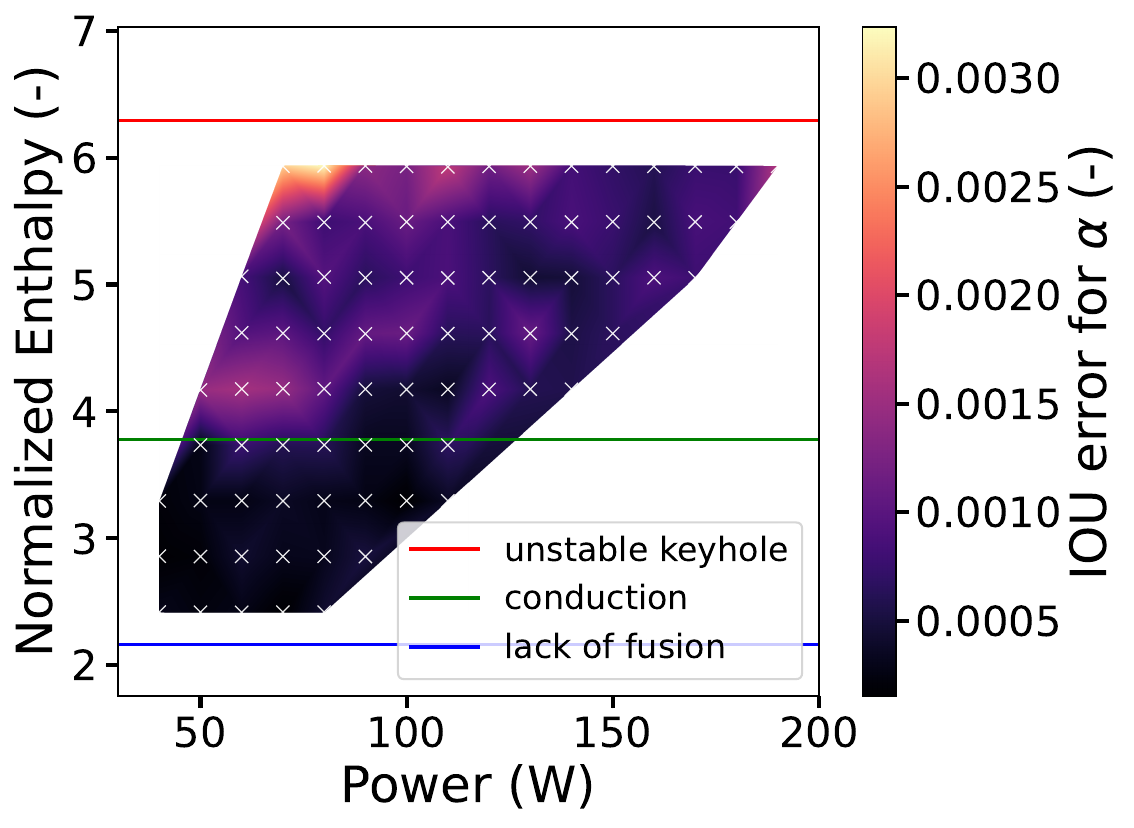}
        \caption{IoU error ($1-$IoU) for $\alpha$ (P--$H$)}
        \label{fig:alpha_iou_ph}
    \end{subfigure}
    \hfill
    \begin{subfigure}[b]{0.48\textwidth}
        \centering
        \includegraphics[width=\textwidth]{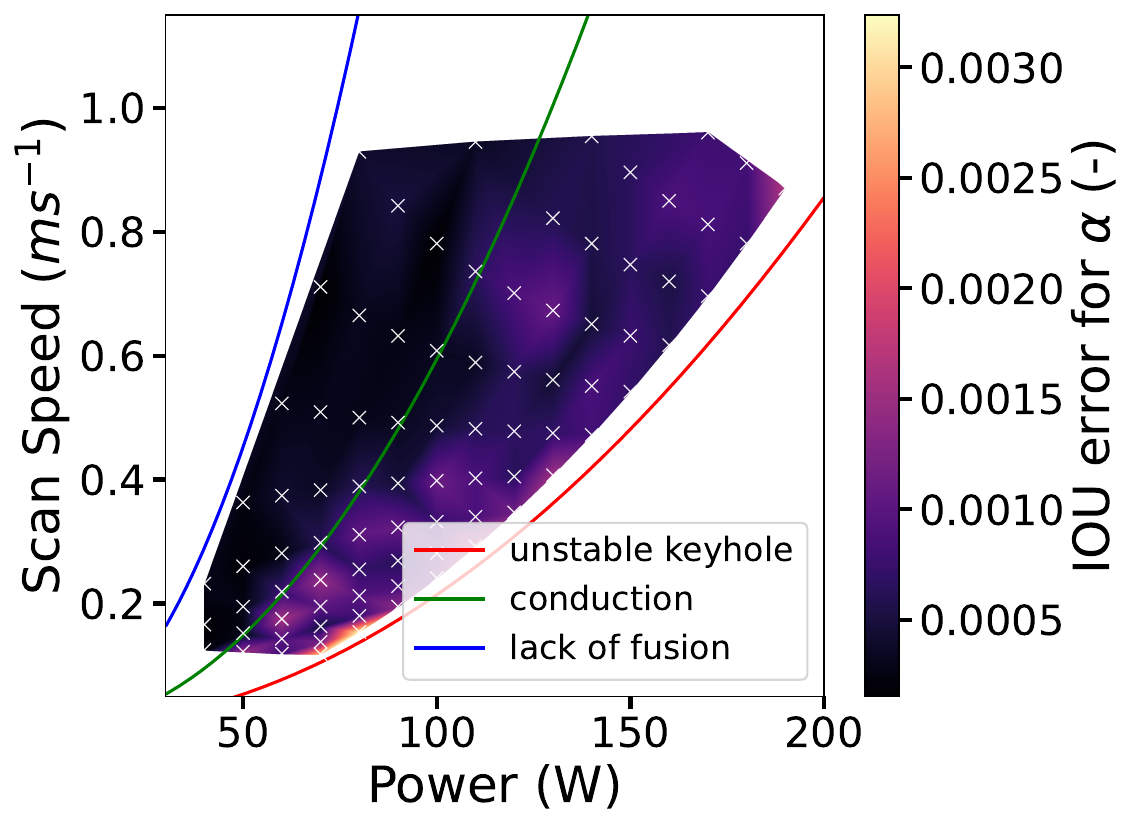}
        \caption{IoU error ($1-$IoU) for $\alpha$ (P--$V_{\mathrm{scan}}$)}
        \label{fig:alpha_iou_pv}
    \end{subfigure}

    \caption{Metal--gas interface ($\alpha$) segmentation performance across the process window. Top row: average absolute error. Bottom row: IoU error ($1-$IoU) computed after thresholding at $\alpha \ge 0.5$. Results are shown in two parameterizations: $(P,H)$ (left) and $(P,V_{\mathrm{scan}})$ (right)}
    \label{fig:alpha_segmentation_2x2}
\end{figure}

\begin{figure}[h]
    \centering
    \begin{subfigure}[b]{0.48\textwidth}
        \centering
        \includegraphics[width=\linewidth]{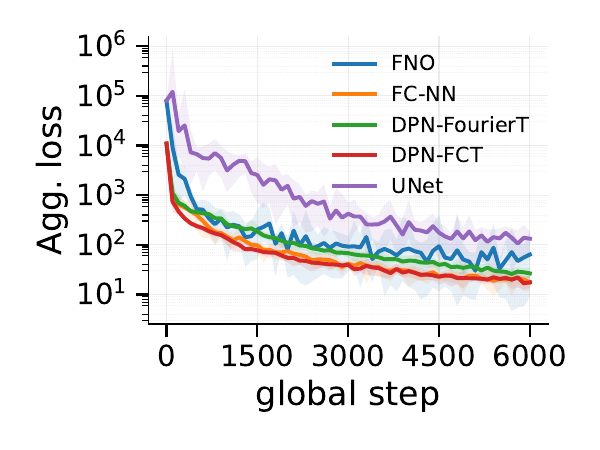}
        \caption{Aggregated training loss}
        \label{fig:fno_train_loss_T}
    \end{subfigure}
    \hfill
    \begin{subfigure}[b]{0.48\textwidth}
        \centering
        \includegraphics[width=\linewidth]{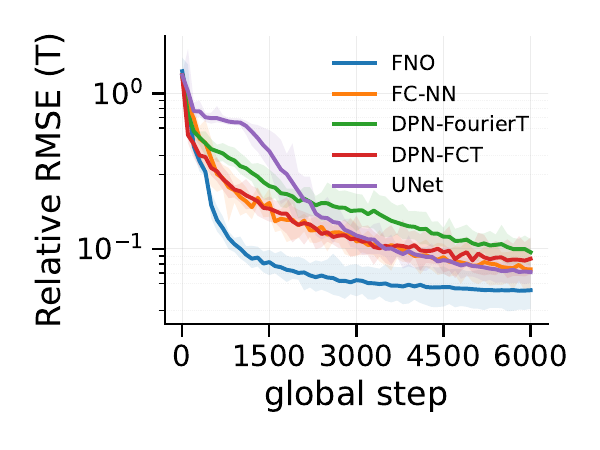}
        \caption{Temperature  Rel.\ RMSE}
        \label{fig:fno_val_relrmse_T}
    \end{subfigure}
    
    \vspace{2mm}

    \begin{subfigure}[b]{0.48\textwidth}
        \centering
        \includegraphics[width=\linewidth]{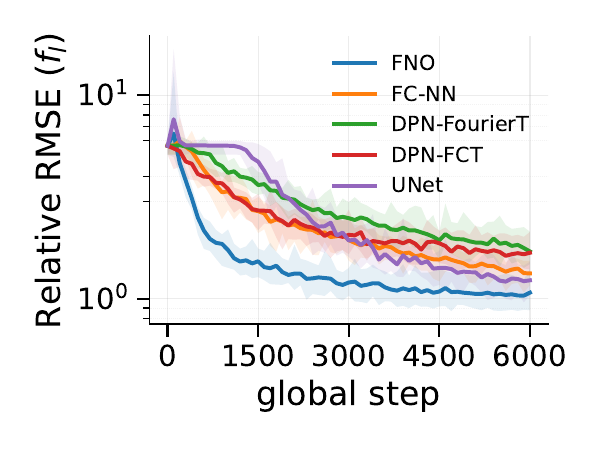}
        \caption{Liquid fraction ($f_\ell$) Rel.\ RMSE}
        \label{fig:fno_train_loss_fl}
    \end{subfigure}
    \hfill
    \begin{subfigure}[b]{0.48\textwidth}
        \centering
        \includegraphics[width=\linewidth]{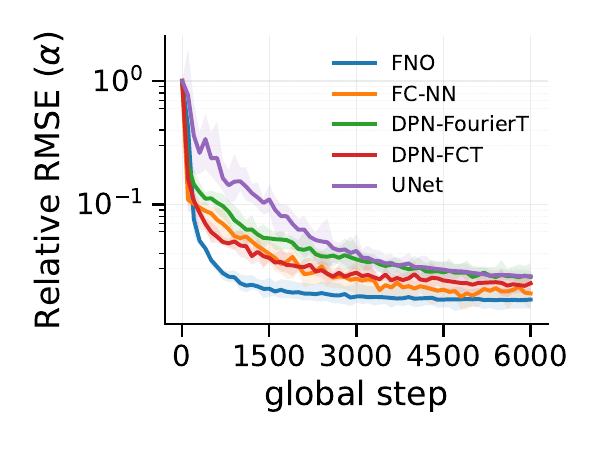}
        \caption{Metal Volume Fraction ($\alpha$) Rel.\ RMSE}
        \label{fig:fno_val_relrmse_fl}
    \end{subfigure}

    \caption{Training convergence metrics for tested architectures under 8-fold cross-validation, excluding the validation-set simulations used for hyperparameter tuning. Consistent convergence is observed across all outputs, with LP-FNO showing the fastest overall convergence among the tested models. The aggregated training losses in (a) of the pointwise models, namely FC-NN and DeepONet, are not directly comparable to the grid-based models because their losses are evaluated on batches of sampled points rather than on the full simulation grid. Therefore, the relative downward trend of each curve is more informative than the absolute aggregated loss value.}
    \label{fig:fno_training_validation_2x2}
\end{figure}

\begin{table*}[t]
\centering
\caption{Architecture hyperparameters for tested models. All models use the same normalized input variables, $x,y,z,V_{\mathrm{scan}},P,H^*$, and predict the same output fields, $T$ and $\alpha$; the liquid fraction $f_\ell$ is computed from the predicted temperature field using \autoref{eq:fl}. Parameters that are not applicable to a given architecture are marked with ``--''. For training,  all models use the ReLoBRaLo loss aggregator with $(\alpha,\beta,\tau)=(0.95,0.99,3.0)$, the Lion optimizer, exponential learning-rate decay with decay steps of 100, 6000 training steps, a moving reference frame, a time-averaging window of 30 steps.}
\label{tab:model_arch_hparams}
\footnotesize
\setlength{\tabcolsep}{4pt}
\renewcommand{\arraystretch}{1.15}
\begin{tabular}{p{0.18\textwidth} p{0.12\textwidth} p{0.15\textwidth} p{0.17\textwidth} p{0.14\textwidth} p{0.11\textwidth}}
\toprule
\textbf{Parameter} 
& \textbf{LP-FNO} 
& \textbf{DPN-FCT} 
& \textbf{DPN-FourierT} 
& \textbf{U-Net} 
& \textbf{FC-NN} \\
\midrule
Input representation 
& grid 
& pointwise branch--trunk 
& pointwise branch--trunk 
& grid 
& pointwise \\

Main architecture 
& FNO  
& DeepONet 
& DeepONet 
& Pix2Pix U-Net 
& FC-NN \\

Spatial dimension 
& 3 
& 3 
& 3 
& 3 
& 3 \\

FNO layers 
& 3 
& -- 
& -- 
& -- 
& -- \\

Fourier modes 
& $[20,18,8]$ 
& -- 
& -- 
& -- 
& -- \\

FNO padding 
& 9 
& -- 
& -- 
& -- 
& -- \\

Decoder / network depth 
& 3 
& 32 / 32 
& 32 / 32 
& 3 downsampling, 3 blocks 
& 48 \\

Hidden width / channels 
& 90 
& 1024 / 1024 
& 1024 / 1024 
& 128 
& 2048 \\

Trunk type 
& -- 
& fully connected 
& Fourier features 
& -- 
& -- \\

Fourier trunk frequencies 
& -- 
& -- 
& axis, $0$--$9$ 
& -- 
& -- \\

Activation function 
& GELU 
& GELU 
& GELU 
& -- 
& GELU \\

Skip connections 
& disabled 
& enabled 
& enabled 
& U-Net connections 
& enabled \\

Total parameter count
& 70,801,900 
& 67,250,176 
& 67,311,616 
& 207,388,418 
& 197,345,282 \\

\midrule
Learning rate 
& \num{1e-4} 
& \num{3e-5} 
& \num{3e-5} 
& \num{1e-5} 
& \num{1e-5} \\

Decay rate 
& 0.96 
& 0.98 
& 0.98 
& 0.95 
& 0.98 \\

Training batch size 
& 1 
& 10000 
& 10000 
& 1 
& 10000 \\

Maximum training steps 
& 6000 
& 6000 
& 6000 
& 6000 
& 6000 \\

Loss aggregator 
& ReLoBRaLo 
& ReLoBRaLo 
& ReLoBRaLo 
& ReLoBRaLo 
& ReLoBRaLo \\

Optimizer 
& Lion 
& Lion 
& Lion 
& Lion 
& Lion \\
\bottomrule

\end{tabular}
\end{table*}

\end{appendices}

\end{document}